\documentclass{article}

\PassOptionsToPackage{numbers}{natbib}
\usepackage[main,final]{neurips_2026}

\usepackage[utf8]{inputenc}
\usepackage[T1]{fontenc}
\usepackage{hyperref}
\usepackage{url}
\usepackage{booktabs}
\usepackage{amsfonts}
\usepackage{amsmath}
\usepackage{nicefrac}
\usepackage{microtype}
\usepackage{xcolor}
\usepackage{graphicx}
\usepackage{subcaption}
\usepackage{multirow}
\usepackage{enumitem}
\usepackage{wrapfig}

\title{Why Speculative Decoding Works Better Than Predicted on Sparse MoE Models}

\author{%
  Ekagra Ranjan \\
  Cohere \\
  \texttt{ekagra@cohere.com} \\
  \And
  Komal K.~Teru \\
  Cohere \\
  \texttt{komal@cohere.com} \\
  \And
  Bharat Venkitesh \\
  Cohere \\
  \texttt{bharat@cohere.com} \\
  \And
  Acyr Locatelli \\
  Cohere \\
  \texttt{acyr@cohere.com} \\
}

\begin{document}

\maketitle

\begin{abstract}
Speculative decoding (SD) accelerates autoregressive inference by leveraging the memory bound nature of decoding, verifying multiple draft tokens in a single forward pass.
For mixture-of-experts (MoE) models, prior work predicts a concave speedup-batch size relationship, with gains peaking at intermediate batch sizes due to expert saturation.
We evaluate this across 10 sparse MoE models (30B--1T parameters, 8--512 experts) and find recent fine-grained models break this trend, showing higher speedup at low batch size.
We find verification cost is governed not only by activated-expert bytes, but also by temporal overlap across draft tokens and GPU execution dynamics.
At low batch size, MoE layers in fine-grained models account for a much smaller share of runtime than of weights; on H100, kernel profiling attributes this mainly to higher bandwidth utilization of MoE kernels relative to attention, with L2 absorbing repeated reads of co-routed experts.
Across draft tokens, routing overlap further reduces the unique experts activated during verification relative to independence assumptions.
We model this effect using a maximum-entropy Iterative
Proportional Fitting (IPF) framework that predicts unique expert counts from routing-overlap statistics with less than 4\% error, and substantially narrows the gap to real-model curves and captures the qualitative low-batch trend from synthetic routing.
\end{abstract}

\section{Introduction}
\label{sec:intro}

Sparse MoE architectures are increasingly used in high-capacity LLMs because they scale model capacity without proportionally increasing per-token compute~\citep{jiang2024mixtral, deepseekai2024deepseekv3, yang2025qwen3, kimik25}. At the same time, long-horizon agentic workloads, test-time scaling, RL rollouts with long-tailed completions, and smaller memory budgets increasingly push inference into low-batch regimes where token latency matters most~\citep{kwon2023vllm, sadhukhan2024magicdec, kariyappa2026sidequest, sun2025contextfolding, snell2025scaling, deepseek2025r1, kimi2025k15, ulorl2025}.
Speculative decoding (SD) is a lossless way to reduce autoregressive latency: a draft model proposes $K$ tokens, and the target model evaluates the corresponding $K{+}1$ positions in a single forward pass, amortizing the cost of a target-model invocation across positions~\citep{leviathan2023fast, chen2023accelerating}. Prior work suggests that SD is less effective for sparse MoE targets in this regime because verification may activate many additional experts and erode the expected gains~\citep{huang2025moesd,li2024eagle,saxena2025utility}.

For a fixed draft model, SD speedup is governed by the acceptance length and the verification cost. Most prior work focuses on improving acceptance length \citep{li2024eagle, li2025eagle3, cai2024medusa}, while verification cost is often treated as a simple function of architecture and batch size \citep{sadhukhan2024magicdec}. For sparse MoEs, however, verification cost depends on how many unique experts are activated across the $B(K{+}1)$ verification positions. MoESD~\citep{huang2025moesd} models this cost under an \emph{activated-expert-bytes} view: in a memory-bandwidth-bound regime~\citep{williams2009roofline}, cost scales with the total expert-weight bytes loaded during verification, approximated as the unique-expert count multiplied by the per-expert byte size. For the MoE models studied there, this view aligns with the expected concave speedup curve and limited low-batch gains.

We ask whether that conclusion continues to hold for recent fine-grained sparse MoE models on modern hardware, and evaluate it across 10 models spanning 30B--1T parameters and 8--512 experts.
For each model, we use the supported SD mechanism (EAGLE, EAGLE-3, MTP, or a separate draft model; details in Appendix~\ref{app:reproducibility}). Most of these models do not exhibit the predicted concave curve. Instead, many remain unusually efficient at low batch sizes and peak at or near batch size~1, precisely where the activated-expert-bytes view predicts limited SD gains for MoE targets.

This discrepancy points to two mechanisms missing from the activated-expert-bytes view:

\begin{enumerate}[nosep, leftmargin=*]
  \item \textbf{Temporal overlap in MoE routing across verification positions.} In fine-grained MoE models~\citep{dai2024deepseekmoe,ludziejewski2024scaling}, adjacent positions share 20–45\% of their activated experts, far above the 4–12\% predicted by an independence baseline (which assumes experts are selected independently across positions given routing probabilities). This temporal overlap reduces the unique expert count during $K{+}1$-position verification by 21--34\% relative to the independence baseline, and accounts for most of the overlap beyond the uniform baseline (which assumes experts are chosen uniformly at random across the expert pool; 69--92\%) in high-sparsity models.

  \item \textbf{GPU execution dynamics.} The activated-expert-bytes view ignores hardware-level effects that decouple parameter size from runtime, including memory hierarchy behavior (e.g., L2 cache reuse), kernel efficiency, and achieved FLOPs/bandwidth. For example, at batch size~1, activated-expert bytes are a poor proxy for latency: for Qwen3-30B FP8 on H100, MoE kernels reach 45--49\% of peak HBM bandwidth while attention kernels reach only 22--25\%, so MoE layers account for only 33\% of forward-pass time despite representing 67\% of model weights. When co-routed positions re-read the same small expert shard, L2 serves the repeated reads: in the same setting, the MoE L2 hit rate rises from 5.72\% in AR decoding to 27.79\% under SD verification. The low MoE runtime share also holds on B200 (35\%), although the L2 hit rate there does not rise under SD: the specific mechanisms depend on the hardware and kernel, but the failure of activated-expert bytes as a cost proxy persists.
\end{enumerate}

Taken together, these results show that SD performance in sparse MoE models is governed not only by \emph{activated-expert bytes}, but also by temporal routing overlap and GPU execution dynamics. This predictive capability is useful in practice because evaluating SD-relevant target and draft architecture choices by direct training-and-benchmarking is expensive, especially at frontier scales. We therefore introduce a maximum-entropy framework based on Iterative Proportional Fitting (IPF) that uses empirical pairwise overlap statistics to predict unique expert counts within 4\% error and reproduce observed speedup curves under synthetic routing. The same framework enables train-free screening of target and draft design tradeoffs before full training, including sparser targets and MoE-versus-dense draft models.

\paragraph{Contributions.}
\begin{itemize}[nosep, leftmargin=*]
  \item We provide a comprehensive empirical study of SD on 10 sparse MoE models spanning 30B--1T parameters and 8--512 experts, and show that fine-grained MoEs remain unusually efficient at low batch sizes and might even peak near batch size~1, contrary to the concave-curve prediction of the activated-expert-bytes view.
  \item We quantify temporal routing overlap across draft positions, tasks, and languages, and show that consecutive positions in fine-grained MoE models reuse substantially more experts than an independent-routing baseline predicts. This overlap persists for rejected draft tokens and reduces the unique expert count during SD verification (\S\ref{sec:temporal}).
  \item We show that \emph{activated-expert bytes} are a poor proxy for runtime at batch size~1 on both H100 and B200, and use detailed GPU profiling to identify the H100 mechanisms, HBM-bandwidth utilization gaps between MoE and attention kernels and L2 cache reuse for small expert shards, and to show that these mechanisms are kernel-dependent (\S\ref{sec:gpu-execution-dynamics}).
  \item We introduce a maximum-entropy IPF framework that predicts unique expert counts within 4\% error and generates realistic token-to-expert assignments. Synthetic-routing experiments reproduce the observed speedup curve, and the same framework enables train-free screening of target and draft architecture tradeoffs, including sparser targets and MoE-versus-dense draft layers, before additional training (\S\ref{sec:end-to-end-validation-via-synthetic-routing}, Appendix~\ref{app:moe-draft-896}).

\end{itemize}

\section{Related Work}
\label{sec:related}

\paragraph{MoE architectures and routing locality.}
Sparse MoE architectures trace back to sparsely-gated MoE~\citep{shazeer2017outrageously} and later scaling and routing designs such as GShard~\citep{lepikhin2021gshard}, BASE Layers~\citep{lewis2021base}, Switch Transformers~\citep{fedus2022switch}, GLaM~\citep{du2022glam}, and Expert Choice Routing~\citep{zhou2022expert}. More recent language-model variants increasingly use finer-grained and shared-expert designs~\citep{komatsuzaki2023upcycling, dai2024deepseekmoe, jiang2024mixtral, deepseekai2024deepseekv3, yang2025qwen3, kimik25}.
Prior work~\citep{olson2025probing, bershatsky2025spatial} shows that expert co-activation is influenced by semantic and positional similarity, and ExpertFlow~\citep{he2024expertflow} exploits this locality for memory-constrained deployment.
We build on this line of work by quantifying how temporal locality across consecutive draft tokens changes speculative-verification cost.

\paragraph{Speculative decoding.}
Early work~\citep{stern2018blockwise} introduced a prefix-validation approach for parallel autoregressive decoding. Modern speculative decoding methods differ mainly in how they construct and verify proposals: draft-model and seq2seq variants~\citep{leviathan2023fast, chen2023accelerating, xia2023specdec}, retrieval- or tree-based proposals~\citep{he2024rest, sun2023spectr, miao2024specinfer, svirschevski2024specexec}, extra decoding heads and multi-token prediction~\citep{cai2024medusa, ankner2024hydra, gloeckle2024multitoken, deepseekai2024deepseekv3, yang2025qwen3}, and higher-capacity or parallel drafting variants~\citep{li2024eagle, li2024eagle2, li2025eagle3, fu2024lookahead, jiang2025scylla, huang2026peagle}.

Recent work has shown that SD can remain effective outside the traditional small-batch dense regime, but for different reasons. 
MagicDec~\citep{sadhukhan2024magicdec} shows that in long-context serving, large batches can again favor SD because KV-cache traffic dominates verification cost.
The closest prior work for MoE is MoESD~\citep{huang2025moesd}, which models verification under an activated-expert-bytes view and accordingly predicts concave speedup curves with limited gains at low batch sizes.
Because memory constraints typically force batch size down as sequence length grows, this low-batch regime is increasingly relevant for the longer-context, long-tailed RL rollout, and agentic workloads emerging in practice.
Our work revisits this question for modern sparse MoEs and identifies a distinct regime in which fine-grained MoEs remain effective even at low batch sizes: temporal routing overlap and GPU execution dynamics yield higher speedup than the activated-expert-bytes view predicts.


\section{Background and Experimental Setup}
\label{sec:background}

Each SD iteration has two phases: (1) a draft model proposes $K$ tokens, and (2) the target model verifies these draft tokens together with the next-token continuation in a single pass over $K{+}1$ positions. SD helps Time per output token (TPOT) only when the savings from fewer target-model calls exceed the overhead of verification—i.e., when acceptance is high and verification is cheap. In MoE models, this verification cost depends not only on model size but also on routing behavior, because verifying multiple tokens may activate overlapping sets of experts across the $K{+}1$ positions.

Because different SD mechanisms have different acceptance behavior, we isolate the target-side verification cost using \emph{target efficiency} (TE)~\citep{huang2025moesd},
\[
\mathrm{TE} = T_{\mathrm{target}}^{\mathrm{AR}} /
T_{\mathrm{target}}^{\mathrm{SD}},
\]
where \(T_{\mathrm{target}}^{\mathrm{AR}}\) denotes the latency of a single-token autoregressive (AR) decoding step on the target model and \(T_{\mathrm{target}}^{\mathrm{SD}}\) denotes the latency of the target model's verification pass over \(K{+}1\) positions in speculative decoding (SD).
A higher TE indicates that the SD verification pass is closer in cost to a standard AR decoding step.
Because TE is independent of acceptance length, it lets us compare target-side verification efficiency across draft mechanisms and model architectures.

The 10 evaluated MoE models span sparsity ratios (k/N, top-k over total experts) from 0.25 (Mixtral) to 0.020 (Qwen3-Next-80B), and per-expert weights from 4.7 MB to 176 MB in FP8, a 37× range that proves critical for GPU execution dynamics (\S\ref{sec:weight-model}); Appendix Table~\ref{tab:models} lists the full architectural details and SD methods.
All experiments use $K{=}3$ draft tokens, vLLM v1 (commit \texttt{7c080dd3c})~\citep{kwon2023vllm}, and MT-Bench~\citep{zheng2023mtbench} as the base dataset; Appendix~\ref{app:reproducibility} provides details on the sweep schedule, token output lengths, and disabled prefix caching, Section~\ref{sec:k-sweep} and Section~\ref{sec:long-context} evaluate larger $K$ and long-context serving, and Appendix~\ref{sec:limitations} discusses the remaining scope of the $K$ and hardware assumptions. All models run on NVIDIA H100 SXM GPUs except for Kimi-K2.5 and DeepSeek-V3 which use NVIDIA B200 GPUs.

\section{Speedup Curve Regimes}
\label{sec:taxonomy}


\begin{figure}[t]
  \centering
  \includegraphics[width=\textwidth]{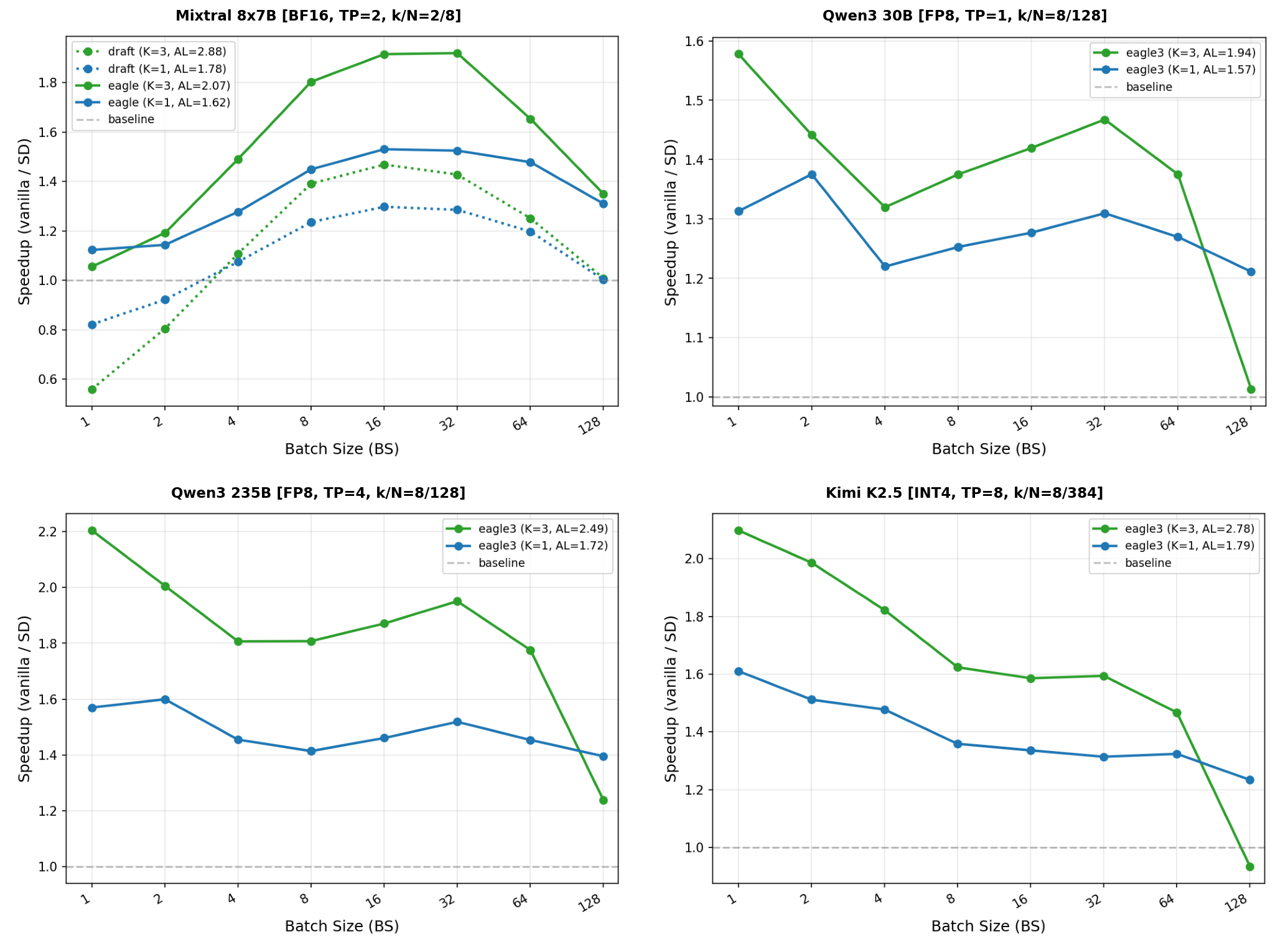}
  \caption{TPOT speedup versus batch size for four representative MoE models.
  Mixtral-8x7B exhibits the predicted concave curve, whereas the other three are non-concave and peak at or near batch size~1.
  }
  \label{fig:speedup-curves}
\end{figure}

We begin by characterizing the observed speedup regimes.
Figure~\ref{fig:speedup-curves} shows TPOT speedup ($\text{TPOT}_\text{vanilla}/\text{TPOT}_\text{sd}$) versus batch size for four representative models, where TPOT denotes wall-clock time per output token. 
Appendix~\ref{app:rfwd} shows all the 10-grid. MoESD~\citep{huang2025moesd} predicts a concave speedup curve that peaks at intermediate batch sizes, where expert activation saturates. 
Under this view, speedup at low batch sizes remains limited because verification must load additional experts.
We observe a \textbf{concave} pattern in less sparse MoEs with larger per-expert weights (e.g., Qwen2-57B, Mixtral-8x7B).
In contrast, sparser models with smaller experts (e.g., Qwen3-235B, Kimi-K2.5) exhibit \textbf{non-concave} curves, often peaking at batch size~1.
The non-concave behavior is also not specific to MT-Bench: Appendix~\ref{app:multi-dataset-speedup} shows that Qwen3-30B and Qwen3-235B retain qualitatively similar speedup curves across multiple datasets (Figure~\ref{fig:multi-dataset-speedup}).



We find that the speedup-curve shape closely tracks target efficiency at batch size~1, which measures how expensive it is for the target model to verify $K{+}1$ tokens instead of decoding a single token. Lower target efficiency leaves more headroom for the classic concave rise, whereas higher target efficiency leads to non-concave behavior, as shown in Figure~\ref{fig:mega-grid}.
Qwen3.5-35B (bf16, TP2) is a notable exception: it exhibits non-concave behavior despite having lower target efficiency than GLM-4.7-Flash (bf16, TP1).
Table~\ref{tab:weight-gap} reports the measured target efficiency for the main-text configurations alongside the activated-expert-bytes prediction; Appendix Table~\ref{tab:target-efficiency-appendix} reports the full cross-model measurements across SD methods, TP settings, and precisions.
We find that target efficiency varies substantially even within a single model family as we change SD method, TP, and precision.
In general, increasing TP or using lower precision increases target efficiency at batch size~1 and shift peak speedup toward smaller batch sizes.
We analyze these effects in more detail in \S\ref{sec:weight-model}.

\section{Limits of the Activated-Expert-Bytes View}
\label{sec:weight-model}


At low batch size $\sim$ 1, decoding processes a single token per sequence, making the linear layers largely memory-bandwidth-bound in the roofline sense~\citep{williams2009roofline}: execution time is therefore dominated by loading weights from HBM rather than by compute.
A first-principles estimate therefore assumes GPU time is proportional to weight bytes:
\begin{equation}
  \text{Target efficiency} \;\approx\; \frac{C_\text{dense} + k \cdot C_\text{expert}}{C_\text{dense} + \text{unique}_\text{sd} \cdot C_\text{expert}},
  \label{eq:weight-model}
\end{equation}
where $C_\text{dense}$ is the per-GPU dense weight bytes (attention + shared expert), $C_\text{expert}$ is the per-expert weight bytes, and $\text{unique}_\text{sd}$ is the empirical unique expert count during SD verification.
Even when using the empirical $\text{unique}_\text{sd}$ measured from actual MT-Bench routing, this activated-expert-bytes model \emph{systematically underestimates} target efficiency (Table~\ref{tab:weight-gap}).
The gap ranges from $+0.06$ (Mixtral, large experts) to $+0.37$ (Qwen3-235B, small experts).
Weight bytes alone do not explain the GPU execution time at batch size~1.
This activated-expert-bytes estimate omits critical model behavior and hardware interactions which we will discuss in detail in the following sections.

\begin{table}[t]
  \caption{TE predicted by the activated-expert-bytes model vs. measured TE at batch size~1 using per-GPU weights.}
  \label{tab:weight-gap}
  \centering
  \small
  \resizebox{\linewidth}{!}{%
  \begin{tabular}{lccccccc}
    \toprule
    Model & $C_\text{expert}$ (MB/GPU) & $C_\text{dense}$ (MB/GPU) & MoE wt\% & $\text{unique}_\text{sd}$ & \text{TE. pred.} & \text{TE observed} & Gap \\
    \midrule
    Mixtral-8x7B (bf16, TP2) & 168.0 & 40.0 & 89.4 & 4.99  & 0.428 & 0.505 & +0.08 \\
    Mixtral-8x7B (fp8, TP1)  & 168.0 & 40.0 & 89.4 & 4.99  & 0.428 & 0.488 & \textbf{+0.06} \\
    Qwen3-30B (bf16, TP1)    & 9.0   & 36.0 & 66.7 & 19.49 & 0.511 & 0.630 & +0.12 \\
    Qwen3-30B (bf16, TP2)    & 4.5   & 18.0 & 66.7 & 19.49 & 0.511 & 0.697 & +0.19 \\
    Qwen3-30B (fp8, TP1)     & 4.5   & 18.0 & 66.7 & 19.49 & 0.511 & 0.818 & \textbf{+0.31} \\
    Qwen3-30B (fp8, TP2)     & 2.25  & 9.0  & 66.7 & 19.49 & 0.511 & 0.860 & +0.35 \\
    Qwen3-235B (bf16, TP8)   & 4.5   & 18.0 & 66.7 & 19.01 & 0.522 & 0.786 & +0.26 \\
    Qwen3-235B (fp8, TP4)    & 4.5   & 17.0 & 67.9 & 19.01 & 0.517 & 0.889 & +0.37 \\
    \bottomrule
  \end{tabular}%
  }
\end{table}

\section{Temporal Overlap in MoE Expert Routing}
\label{sec:temporal}



To explain why verification cost stays low, we analyze expert overlap across consecutive tokens using three levels that isolate the contribution of temporal overlap; formal definitions and derivations are given in Appendix~\ref{app:overlap-formulas}:
\begin{enumerate}[nosep,leftmargin=*]
  \item \textbf{Uniform baseline} ($k/N$): assumes all experts equally likely and tokens independent.
  \item \textbf{Independence baseline}: uses the empirical per-layer expert probabilities $\{p_e\}$ while assuming tokens are routed independently.
  This captures savings from non-uniform expert popularity.
  \item \textbf{Empirical}: measured from consecutive tokens on a real dataset, preserving temporal overlap.
\end{enumerate}

We report routing-overlap statistics for the models in our SD setup; details in Appendix~\ref{app:routing-inertia}.
Appendix~\ref{app:overlap-decay-full} shows overlap for tokens separated by $j$ positions, where step~$j$ denotes a separation of $j$ tokens.
Across the eight measured models, mean empirical overlap declines from 35.1\% at step~1 to 26.4\% at step~3, while remaining 3.3$\times$, 2.7$\times$, and 2.5$\times$ above the mean independence baseline of 10.5\% at steps 1--3, respectively.

\noindent Table~\ref{tab:overlap} reports step-1 overlap ratios, defined as the fraction of top-$k$ experts shared between adjacent tokens along with its decomposition, for four representative models; a broader model analysis is provided in Appendix~\ref{app:overlap-formulas}, Table~\ref{tab:overlap-full}.
All models shows empirical overlap far above both the uniform baseline and the independence baseline. The temporal overlap beyond the independence baseline dominates in 6 of 8 models, contributing 69--92\% of total overlap.
Mixtral ($k/N{=}0.25$) and Qwen2-57B ($k/N{=}0.125$) are exceptions: their large combinatorial baseline leaves less room for temporal effects.
We defer per-layer patterns to Appendix~\ref{app:per-layer-overlap}.

\begin{table}[t]
  \caption{Step-1 expert overlap decomposition.
  \emph{Temporal} = empirical $-$ independence; \emph{Temporal/Total} = fraction of overlap attributable to temporal overlap.}
  \label{tab:overlap}
  \centering
  \small
  \begin{tabular}{lcccccc}
    \toprule
    Model & $k/N$ & Uniform & Indep & Empirical & Temporal & Temp/Total \\
    \midrule
    Mixtral-8x7B    & 0.250  & 0.250 & 0.253 & 0.355 & 0.102 & 28.7\% \\
    Qwen3-30B       & 0.063  & 0.063 & 0.119 & 0.431 & 0.312 & \textbf{72.4\%} \\
    Qwen3-235B      & 0.063  & 0.063 & 0.098 & 0.447 & 0.349 & \textbf{78.2\%} \\
    Kimi-K2.5       & 0.021  & 0.021 & 0.040 & 0.296 & 0.256 & \textbf{86.4\%} \\
    \bottomrule
  \end{tabular}
\end{table}


\subsection{Applicability to draft tokens.}
\label{sec:draft-temporal}
The overlap measurements above use accepted continuations.
If rejected draft tokens were routed to substantially different experts, accepted-token overlap would mischaracterize the routing patterns encountered during SD verification.
In aggregate MT-Bench, however, rejected tokens share a similar fraction of top-$k$ experts with the preceding token, and in some cases slightly more.
The deconfounded accepted-vs.-first-rejection comparison in Appendix~\ref{app:routing-inertia} shows the same pattern for Qwen3-235B at step~1, where first-rejection tokens have higher overlap than accepted tokens (0.488 vs.\ 0.447; Table~\ref{tab:routing-deconfounded}).
We therefore treat accepted-token overlap as a conservative proxy for aggregate verification cost.

\subsection{Sub-Linear Verification Cost}
\label{sec:verification-cost}

The practical consequence for SD is that verification loads far fewer unique experts than the worst case predicts.
Table~\ref{tab:unique-ipf} compares the per-layer unique expert count when verifying $K{+}1{=}4$ tokens under all three baselines for four representative models.
For Qwen3-30B ($k{=}8$, $N{=}128$), verifying four tokens loads only 2.4$\times k$ unique experts per layer, rather than the 3.6$\times k$ predicted under uniform random routing.
This 33\% reduction directly lowers the MoE forward-pass cost during SD verification.

\begin{table}[t]
  \caption{Unique experts ($K{=}3$) and IPF prediction accuracy.
  Emp = empirical; Unif = uniform.}
  \label{tab:unique-ipf}
  \centering
  \small
  \resizebox{\linewidth}{!}{%
  \begin{tabular}{lcccccccccc}
    \toprule
    & & & \multicolumn{3}{c}{Unique experts (4 tokens)} & & \multicolumn{4}{c}{IPF prediction} \\
    \cmidrule(lr){4-6} \cmidrule(lr){8-11}
    Model & $k$ & $N$ & Unif & Indep & Emp & Emp/$k$ & Uniform IPF & Err & Skewed IPF & Err \\
    \midrule
    Mixtral-8x7B   & 2  & 8   & 5.5  & 5.5  & 5.0  & 2.5$\times$ & 4.99  & +0.1\% & 4.99  & +0.1\% \\
    Qwen2-57B      & 8  & 64  & 26.5 & 26.0 & 24.6 & 3.1$\times$ & 24.72 & +0.5\% & 24.63 & +0.1\% \\
    Qwen3-30B      & 8  & 128 & 29.1 & 26.9 & 19.5 & 2.4$\times$ & 20.08 & +3.1\% & 19.72 & +1.2\% \\
    Kimi-K2.5      & 8  & 384 & 31.0 & 30.2 & 23.3 & 2.9$\times$ & 23.88 & +2.7\% & 23.66 & +1.7\% \\
    \bottomrule
  \end{tabular}%
  }
\end{table}

Two sources contribute to the savings.
The step from the uniform to the independence baseline, which captures expert popularity, is modest: at most 8\% (typically 2--5\%) across the measured models (Table~\ref{tab:unique-ipf-full}).
The step from the independence baseline to the empirical measurement, which captures temporal overlap beyond independent routing, is dominant: a further 21--34\% reduction for the six high-sparsity models ($k/N \le 1/16$).
Models with coarse routing (Mixtral, Qwen2-57B) see smaller temporal savings because the independence baseline is already close to the empirical value.

This property appears largely data-agnostic for Qwen3-30B: across 13 Spec-Bench~\citep{xia2024specbench} task categories, the step-1 overlap ratio varies by only $\pm$0.03 around the MT-Bench value of 0.431, and across seven Aya multilingual splits the full decay curve shifts upward slightly rather than weakening (Appendix~\ref{app:overlap-heldout}).
These results suggest that sub-linear scaling holds across a broad range of realistic workloads.

\subsection{Predicting Unique Experts and Validating the Framework}
\label{sec:ipf}


We develop a maximum-entropy \textbf{Iterative Proportional Fitting (IPF)}~\citep{deming1940leastsquares, csiszar1975idivergence} framework that predicts unique expert counts from pairwise routing-overlap statistics alone, without running inference.
Each expert's activation across $B{=}4$ consecutive tokens is modeled as correlated binary variables $X_1, \ldots, X_4$, with constraints
\begin{equation}
  P_e(X_t{=}1) = p_e, \qquad P_e(X_t{=}1 \wedge X_{t+j}{=}1) = \text{overlap}_j \times p_e \quad (j \in \{1,2,3\}).
\label{eq:ipf-constraints}
\end{equation}
where $p_e$ denotes the empirical probability that expert $e$ is selected on a token (or $k/N$ in the uniform variant), assumed to be position-independent across the four-token window for simplicity, and $P_e$ denotes the fitted joint distribution over that expert's activation pattern across the four-token window.
IPF solves for the maximum-entropy joint distribution over $2^4{=}16$ states, from which the expected unique count follows as $\sum_e[1 - P_e(X_1{=}0,\ldots,X_4{=}0)]$.

We use \emph{uniform} IPF to denote the variant that assumes uniform expert probabilities while matching pairwise overlap across consecutive tokens.
We use \emph{skewed} IPF to denote the variant that uses empirical per-layer expert probabilities together with the same overlap ratios; this is the predictor we use throughout.
Skewed IPF predicts empirical unique-expert counts within 4\% for every measured model (maximum $+3.7\%$, Qwen3-Next-80B), and uniform IPF within 5\% (maximum $+4.7\%$) (Appendix Table~\ref{tab:unique-ipf-full}; Table~\ref{tab:unique-ipf} shows four representative models).
Details of the IPF formulation and synthetic routing generation are in Appendix~\ref{app:synthetic-routing}.


\section{Interaction of GPU execution of sparse MoE on Speculative Decoding}
\label{sec:gpu-execution-dynamics}
The preceding analysis shows that SD verification activates substantially fewer unique experts than an independence-based routing model would predict. 
However, temporal overlap alone does not fully explain the gap between activated-expert-bytes estimates and measured verification cost at batch size~1. 
The realized cost of loading experts also depends on how sparse MoE kernels execute on GPU hardware. 
We therefore turn to GPU execution dynamics, which govern how expert loading translates into observed latency through bandwidth utilization and cache behavior.

We profile Qwen3-30B and Mixtral at batch size~1 on H100 using Nsight Systems (kernel timing) and Nsight Compute (hardware counters), and repeat the profiling for Qwen3-235B on H100 and for Qwen3-30B and Mixtral on B200 (Appendix~\ref{app:profiling-extra}). 
We focus on FP8 variants because they exhibit a larger gap and better expose hardware-level effects.

We summarize the metrics and hardware terms below.
 
\begin{center}
\small
\begin{tabular}{ll|ll}
\multicolumn{2}{c}{\textbf{Metrics}} & \multicolumn{2}{c}{\textbf{Hardware terms}} \\
\hline
SD scaling & \(S = T^\text{SD} / T^\text{vanilla}\) 
& CTA & cooperative thread array (thread block) \\
MoE scaling & \(S_{\text{MoE}} = T^\text{SD}_{\text{MoE}} / T^\text{vanilla}_{\text{MoE}}\) 
& SM & streaming multiprocessor (H100: 132 SMs) \\
Attention scaling & \(S_{\text{Attn}} = T^\text{SD}_{\text{Attn}} / T^\text{vanilla}_{\text{Attn}}\) 
& HBM BW & high-bandwidth memory bandwidth \\
 & 
& L2 hit rate & fraction of accesses served by L2 (H100: 50MB) \\
\end{tabular}
\captionof{table}{Metrics and hardware terms used in profiling.}
\label{tab:metrics-hw}
\end{center}

On H100, Table~\ref{tab:profiling} summarizes the key measurements and reveals two mechanisms behind the gap: small experts (Qwen3-30B) create large disparities in bandwidth utilization and L2 cache behavior between attention and MoE kernels, while large experts (Mixtral) largely eliminate both.

\begin{wraptable}{r}{0.53\linewidth}
  \vspace{-0.8em}
  \centering
  \scriptsize
  \setlength{\tabcolsep}{3pt}
  \caption{GPU profiling comparison at batch size~1 on H100.}
  \label{tab:profiling}
  \resizebox{\linewidth}{!}{%
  \begin{tabular}{lcc}
      \toprule
      Metric & Qwen3-30B & Mixtral-8x7B \\
      \midrule
      Expert size (FP8)        & 4.5\,MB           & 168\,MB \\
      MoE share of weight      & 66.7\%            & 89.4\% \\
      MoE share of GPU time    & \textbf{33\%}     & 75\% \\[3pt]
      Attn CTAs / 132 SMs      & 16~(12\%)         & 96~(73\%) \\
      Attn HBM BW (\% peak)    & 22--25\%          & 62\% \\
      MoE CTAs / 132 SMs       & 96--128           & 132 \\
      MoE HBM BW (\% peak)     & 45--49\%          & 79\% \\[3pt]
      Attn SD scaling $S$      & 1.02              & 1.01 \\
      MoE SD scaling $S$       & 1.57              & 2.19 \\
      L2 hit rate (AR$\to$SD)  & 5.72\%$\to$\textbf{27.79\%} & 1.68\%$\to$1.7\% \\[3pt]
      TE pred.                 & 0.51              & 0.43 \\
      TE observed              & 0.82              & 0.49 \\[3pt]
      Weight model gap         & +0.31             & +0.06 \\
      \bottomrule
    \end{tabular}%
  }
  \vspace{-0.8em}
\end{wraptable}

\paragraph{L2 cache reuse across co-routed experts.}
The activated-expert-bytes model assumes that MoE cost scales linearly with the number of unique experts loaded.
At batch size~1, the default H100 MoE kernel processes each (token, expert) pair in its own thread block, so when multiple verification tokens are routed to the same small expert (Section~\ref{sec:temporal}), its weights are read once per token and the repeated reads are served from L2 rather than HBM.
For Qwen3-30B, the L2 hit rate rises from 5.72\% to 27.79\% for 4.5\,MB experts, whose working set partially fits within the H100's 50\,MB L2 cache.
For Mixtral, the kernel groups co-routed tokens by expert under SD, so each expert is read once and there are no repeated reads to absorb (1.68\% in both modes); its 168\,MB experts would far exceed L2 in any case.
Without L2, these repeated reads would make HBM traffic grow with the number of (token, expert) pairs rather than with the number of unique experts; L2 brings it back close to the unique-expert bytes that the activated-expert-bytes model assumes.
The lower $S_\text{MoE}$ of Qwen3-30B (1.57 vs.\ 2.19 for Mixtral, despite a similar 2.4--2.5$\times$ unique-expert multiple; Section~\ref{sec:temporal}) therefore likely stems mainly from bandwidth utilization, discussed below.

\paragraph{Factors affecting L2 cache reuse.}
Table~\ref{tab:weight-gap} shows that, within a model family, higher tensor parallelism and lower-precision weights tend to increase target efficiency.
Qwen3-30B illustrates the effect concretely: target efficiency rises from 0.630 to 0.697 when TP increases from 1 to 2 at BF16, and from 0.630 to 0.818 when weights are quantized from BF16 to FP8 at TP1; the combined FP8, TP2 configuration reaches 0.860.
One contributing mechanism is that tensor parallelism reduces the per-GPU expert shard size, while quantization further shrinks each shard, allowing more of the co-routed expert working set to stay in L2 between repeated reads.
By contrast, Mixtral's experts are so large that even a single expert does not fit in L2, so the scope for additional reuse is limited and the improvement remains small across the available EAGLE configurations.
Appendix Figure~\ref{fig:qwen3-30b-config-sweep} shows the corresponding Qwen3-30B configuration sweep.

\paragraph{HBM bandwidth utilization disparity.}
At batch size~1, FP8 attention GEMM kernels launch far fewer CTAs than FP8 MoE kernels, leaving many SMs idle and therefore achieving substantially lower HBM bandwidth (Table~\ref{tab:profiling}).
This pattern is consistent with prior work on attention kernels, which emphasizes IO-aware scheduling and improved work partitioning~\citep{dao2022flashattention, dao2024flashattention2}.
As a result, MoE accounts for \emph{less} GPU time than its weight share would suggest.
For Qwen3-30B, the roughly 2$\times$ bandwidth advantage of MoE GEMMs over attention GEMMs compresses the MoE contribution from 67\% of weights to 33\% of runtime.
For Mixtral, MoE and attention GEMMs achieve more similar bandwidth, so the MoE runtime share remains closer to its weight share.
Verifying $K{+}1$ tokens barely increases attention cost ($S{=}1.02$) because KV-cache reads are reused across consecutive draft tokens, making MoE the principal source of SD slowdown in this regime.
Consequently, the smaller MoE runtime share in Qwen3-30B makes SD verification faster than the activated-expert-bytes model would predict.
For Mixtral, larger matrices enable more CTAs for attention, narrowing both the bandwidth gap and the activated-expert-bytes-model error.

\paragraph{Expert size as the unifying factor.}
On H100, both mechanisms are governed primarily by expert size.
Large experts achieve high bandwidth utilization and exceed L2 capacity, so execution time remains approximately proportional to weight bytes and the activated-expert-bytes model is relatively accurate.
Small experts leave bandwidth underutilized, making execution time substantially smaller than the activated-expert-bytes model would predict, while L2 absorbs repeated reads of co-routed experts.
A full three-step correction chain that raises the estimate from 0.51 to 0.872 for Qwen3-30B is provided in Appendix~\ref{app:three-step}.
Qwen3-235B (FP8, TP4) reproduces the same mechanisms on H100, with the largest weight-model gap among the profiled models (+0.38; Appendix~\ref{app:profiling-extra}).


\section{End-to-end validation via synthetic routing.}
\label{sec:end-to-end-validation-via-synthetic-routing}
Section~\ref{sec:temporal} and Section~\ref{sec:gpu-execution-dynamics} show that both temporal overlap and GPU execution dynamics shape the SD speedup curves of MoE models.
We therefore conduct synthetic experiments that combine these two factors to predict the speedup curves without using trained weights.
The only assumption is access to reasonable expert-overlap statistics, obtained either from a model in the same family or from the averages in Table~\ref{tab:overlap-decay-full}.
To evaluate the IPF framework from Section~\ref{sec:ipf} and its synthetic-routing construction from Appendix~\ref{app:synthetic-routing}, we replace the model's learned routing with IPF-generated synthetic routing while keeping all other inference settings fixed (dummy weights, same hardware).
Because this method runs the real kernels on the target hardware, it captures hardware- and kernel-specific execution effects, such as those that differ between H100 and B200, without modeling them explicitly.
We define the \emph{proxy target efficiency} as
$T_\text{vanilla}^\text{proxy} / T_\text{SD}^\text{proxy}$,
where $T_\text{vanilla}^\text{proxy}$ = forward pass + logits + sampling, and $T_\text{SD}^\text{proxy}$ = forward pass + logits + rejection sampling + draft.

\begin{figure}[t]
  \centering
  \captionsetup[subfigure]{font=footnotesize, justification=centering}
  \begin{subfigure}[t]{0.32\textwidth}
    \centering
    \includegraphics[width=\linewidth]{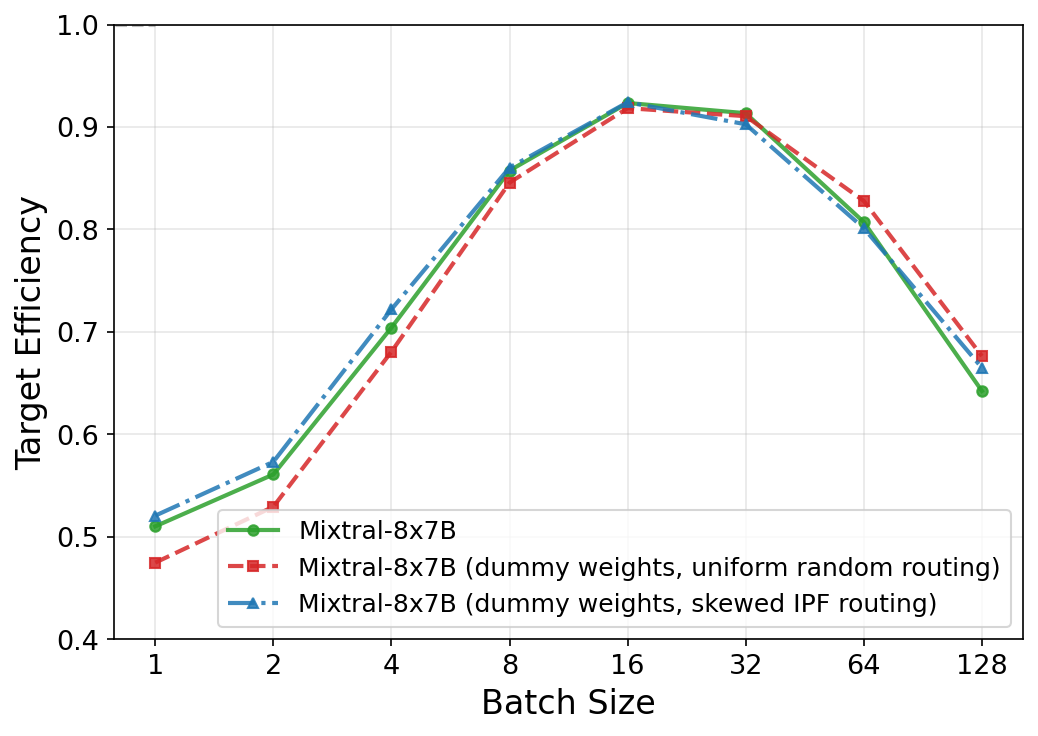}
    \caption{Mixtral-8x7B bf16}
  \end{subfigure}
  \hfill
  \begin{subfigure}[t]{0.32\textwidth}
    \centering
    \includegraphics[width=\linewidth]{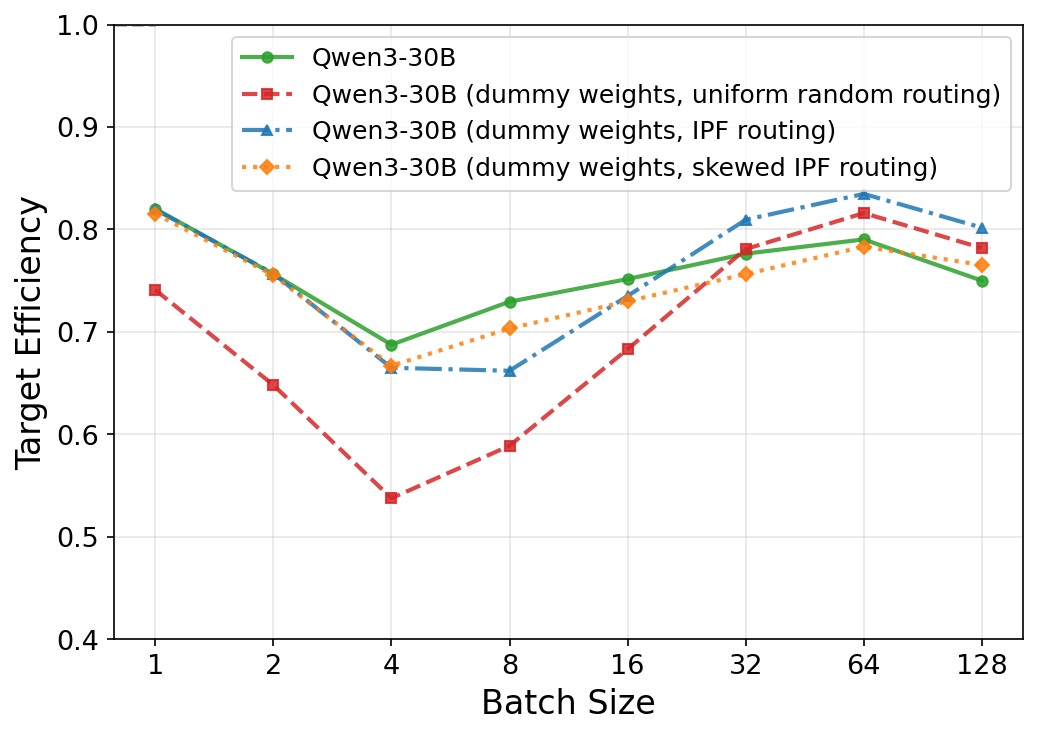}
    \caption{Qwen3-30B fp8}
    \label{fig:dummy-validation-qwen3-30b-tp1}
  \end{subfigure}
  \hfill
  \begin{subfigure}[t]{0.32\textwidth}
    \centering
    \includegraphics[width=\linewidth]{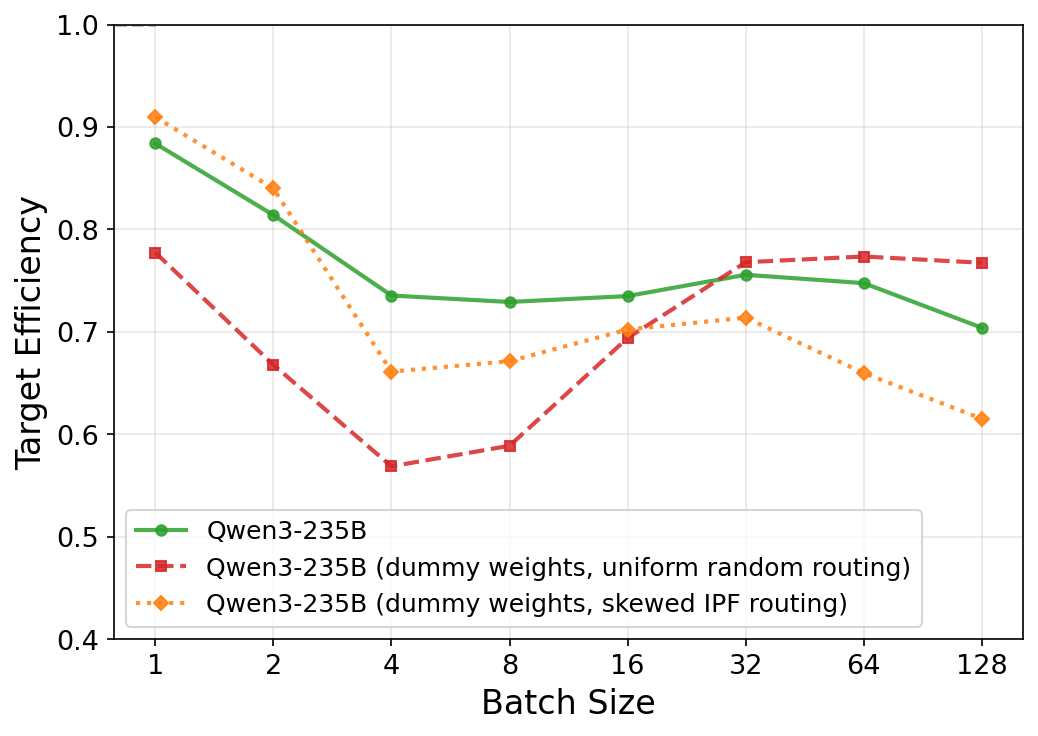}
    \caption{Qwen3-235B fp8}
  \end{subfigure}
  \caption{Target efficiency under different routing strategies.}
  \label{fig:dummy-validation}
\end{figure}

We measure the \emph{proxy target efficiency} under three routing assumptions: uniform random routing, uniform IPF routing with the same routing-overlap statistics in every MoE layer, and per-layer \emph{skewed} IPF routing.
Figure~\ref{fig:dummy-validation} compares these three settings.
The uniform-random baseline consistently underestimates the target efficiency at small batch sizes, and the error is larger for sparser MoE models.
Per-layer IPF synthetic routing closely tracks the real-model curves at small batch sizes and remains closer than uniform random routing across the full batch-size range.
For Qwen3-30B (Figure~\ref{fig:dummy-validation}\subref{fig:dummy-validation-qwen3-30b-tp1}), we show all three routing variants for comparison.
Although the skewed IPF variant is more accurate, the uniform IPF variant still captures the qualitative batch-size trends of the real-model SD speedup curves.

Residual gaps between the IPF-generated synthetic routing and the real-model curves are expected because the synthetic routing does not exactly reproduce the full routing distribution of the real model.
In particular, higher-order statistics not captured by the pairwise routing-overlap constraints and position-independent per-expert marginals in Eq.~\ref{eq:ipf-constraints} may still affect performance.
Nevertheless, predicting these curves without training a model requires only uniform IPF together with routing-overlap statistics taken either from the same model family or from the averages in Table~\ref{tab:overlap-decay-full}.
These results show that routing-overlap statistics captured by the IPF framework substantially reduce the gap between synthetic and real-model curves relative to the uniform-random baseline, while preserving the relevant hardware-execution dynamics.

\section{Discussion}
\label{sec:discussions}

\begin{figure}[t]
  \centering
  \captionsetup[subfigure]{font=footnotesize, justification=centering}
  \begin{subfigure}[t]{0.32\textwidth}
    \centering
    \includegraphics[width=\linewidth]{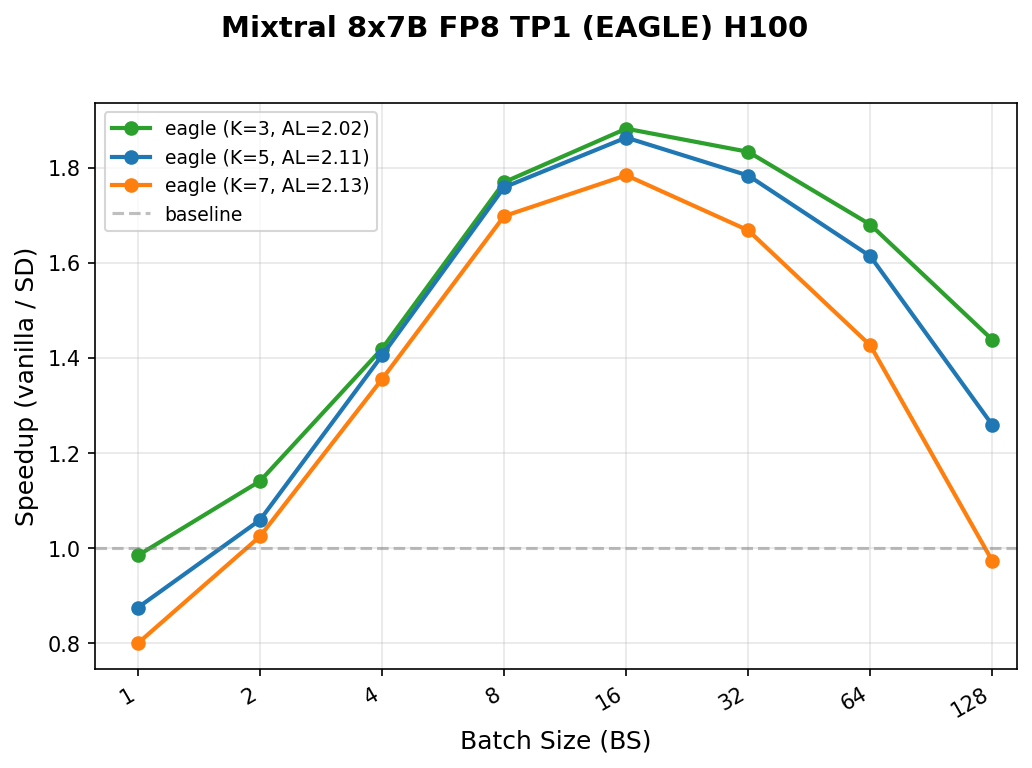}
    \caption{Mixtral-8x7B (FP8, TP1)}
  \end{subfigure}\hfill
  \begin{subfigure}[t]{0.32\textwidth}
    \centering
    \includegraphics[width=\linewidth]{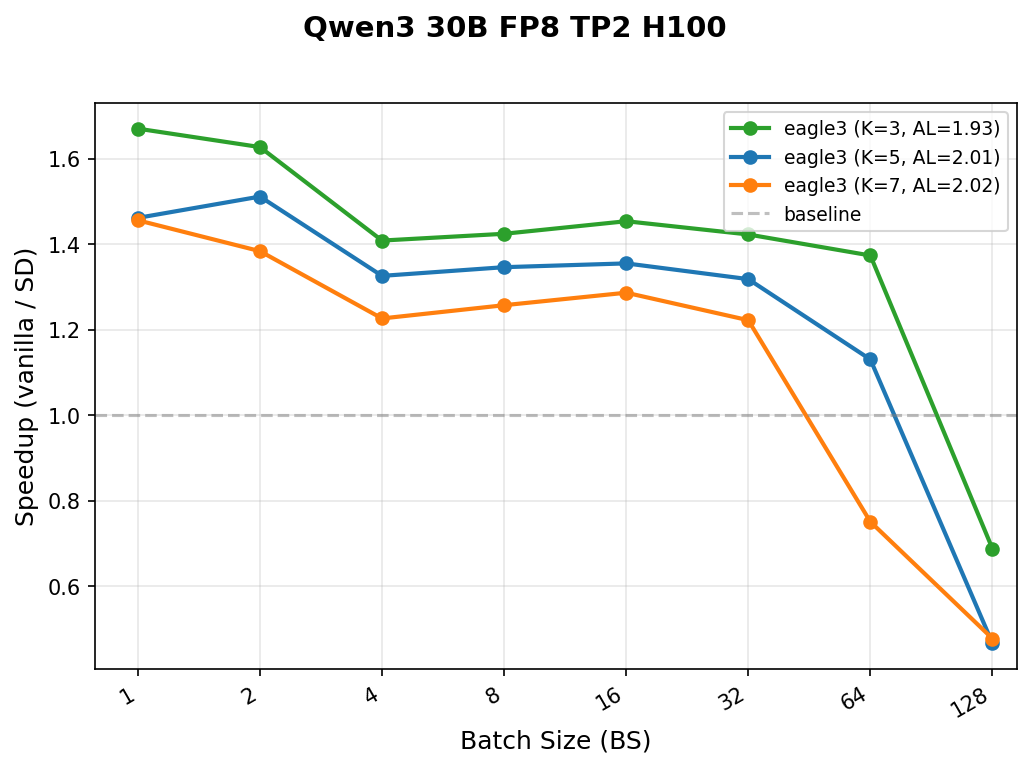}
    \caption{Qwen3-30B (FP8, TP2)}
  \end{subfigure}\hfill
  \begin{subfigure}[t]{0.32\textwidth}
    \centering
    \includegraphics[width=\linewidth]{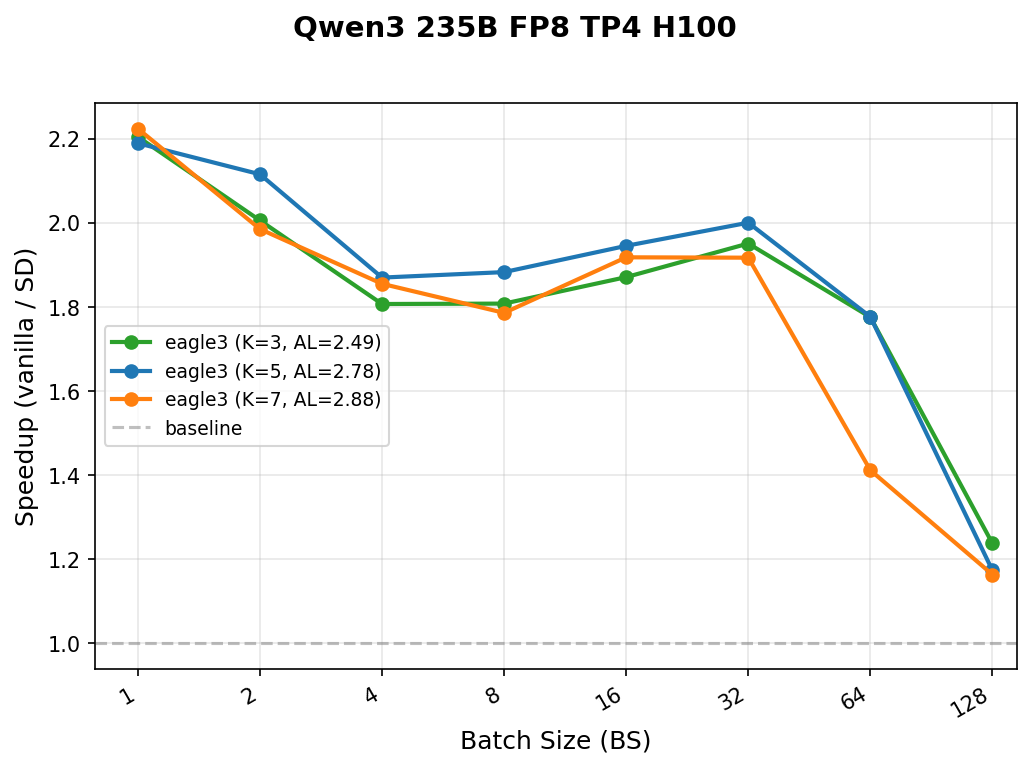}
    \caption{Qwen3-235B (FP8, TP4)}
  \end{subfigure}
  \caption{SD speedup vs.\ batch size for $K{=}3/5/7$ on MT-Bench (H100). Each model keeps
  its curve shape across $K$; larger $K$ mostly shifts the curve down, except for Qwen3-235B,
  whose stronger drafter keeps $K{=}5$ competitive.}
  \label{fig:k-sweep}
\end{figure}

\subsection{Effect of the Speculative Window $K$}
\label{sec:k-sweep}

\begin{wrapfigure}[13]{r}{0.45\textwidth}
  \vspace{-1.2em}
  \centering
  \includegraphics[width=\linewidth]{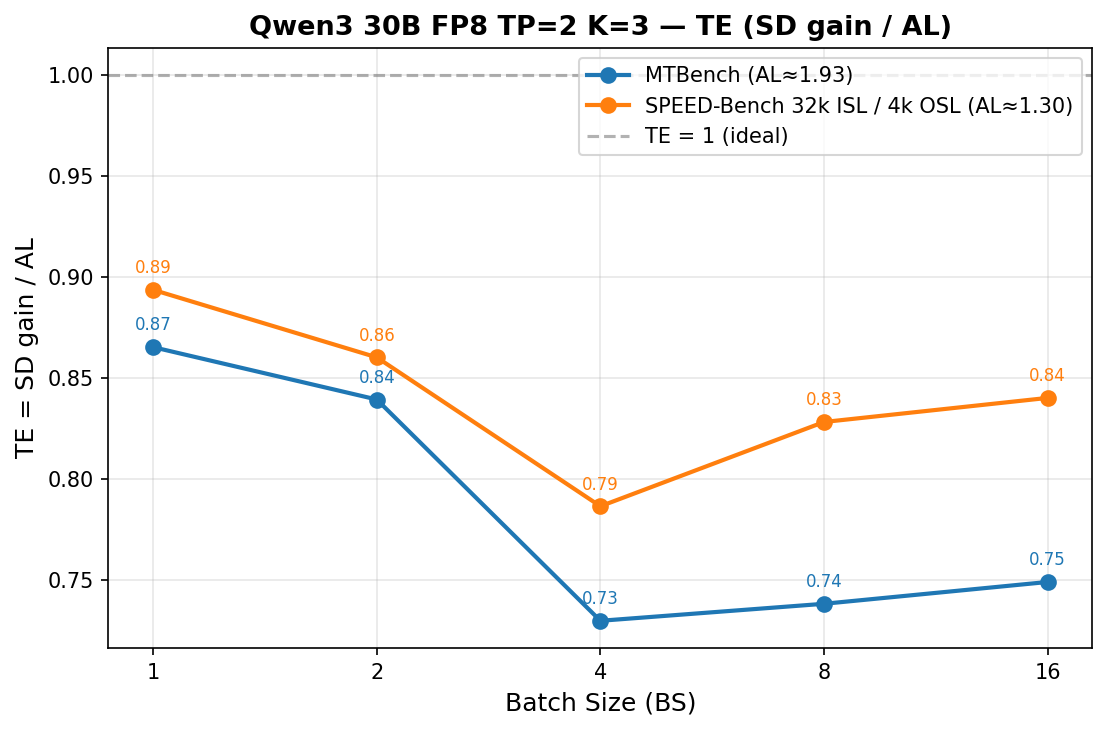}
  \caption{Target efficiency vs.\ batch size for Qwen3-30B (FP8, TP2, EAGLE-3, $K{=}3$) on H100: MT-Bench vs.\ SPEED-Bench 32k.}
  \label{fig:long-context}
  \vspace{-1em}
\end{wrapfigure}
Our main experiments fix $K{=}3$, a common production setting~\citep{tang2025llamasd,nvidia2025gptosseagle3,redhat2026gptossspec}.
To test whether the observed curve shapes depend on this choice, we repeat the end-to-end sweep at $K\in\{3,5,7\}$ for one model from each regime: Mixtral-8x7B (FP8, TP1, EAGLE), Qwen3-30B (FP8, TP2, EAGLE-3), and Qwen3-235B (FP8, TP4, EAGLE-3), all on H100 with MT-Bench (Figure~\ref{fig:k-sweep}).
For all three models, the curve family is the same at every $K$: Qwen3-30B and Qwen3-235B remain non-concave and peak at batch size~1, whereas Mixtral remains concave, with its best speedup near batch size~16.
Our curve taxonomy (\S\ref{sec:taxonomy}) is therefore not an artifact of $K{=}3$.

\subsection{Long-Context Inference}
\label{sec:long-context}

Our main sweeps use short MT-Bench prompts.
To test the long-context regime, where KV-cache traffic becomes increasingly dominant, we repeat the end-to-end sweep for Qwen3-30B (FP8, TP2, EAGLE-3, $K{=}3$) on the \texttt{throughput\_32k} split of SPEED-Bench~\citep{abramovich2026speedbench}, which has prompts of about 32k tokens, with 4096-token outputs (Figure~\ref{fig:long-context}).
The KV cache limits the maximum batch size to 24.
Target efficiency is higher than on MT-Bench at every batch size.
This is consistent with KV-cache reads, which both vanilla decoding and SD verification incur, diluting the relative cost of the extra expert weights loaded during verification.
The curve also remains non-concave: target efficiency is highest at batch size~1, dips at batch size~4, and recovers at larger batch sizes.
Long context thus makes verification relatively cheaper for the MoE target; MagicDec~\citep{sadhukhan2024magicdec} studies this regime in depth.
\subsection{Implications for MoE Draft Layers}
\label{sec:moe-draft}


Current SD drafters commonly use a dense transformer layer.
In Appendix~\ref{app:moe-draft}, we use the framework from Section \ref{sec:end-to-end-validation-via-synthetic-routing} to analyze the effect of an MoE drafter when the target model is also an MoE model.
The analysis suggests that MoE draft layers may become attractive in sufficiently sparse regimes.
At low-to-medium batch sizes, an MoE drafter does not improve target efficiency, but if trained effectively it could still improve end-to-end speedup through longer acceptance lengths enabled by its additional parameters.
At larger batch sizes, the MoE drafter becomes more attractive because it remains less compute-bound than a dense drafter when the target model itself is MoE.
To probe a sparser setting than Qwen3-30B, we also simulate a 896-expert architecture (<1\% activations) and show in Appendix~\ref{app:moe-draft-896} that the MoE drafter becomes more favorable as the target model remains bandwidth-bound.
Because this analysis is synthetic and does not measure acceptance-length gains from trained MoE drafters, we interpret it as a design study rather than direct evidence for trained MoE drafters.

\subsection{Routing Inertia in Rejected Draft Tokens}
\label{sec:routing-inertia-discussion}

Expert routing does not cleanly separate accepted from rejected draft tokens in a way that would support early termination. In the aggregate MT-Bench analysis, fine-grained models show comparable or slightly higher overlap for rejected tokens, and held-out Spec-Bench and Aya~\citep{singh2024aya} slices preserve the same tendency in median, with some task- and language-specific reversals (Appendix~\ref{app:routing-heldout}).
Even in the largest stable aggregate effect (Qwen3-235B, Cohen's $d \approx -0.3$), the accepted and rejected distributions still overlap by ${\sim}85\%$, and most held-out slices remain negligible or small in effect size.
These results suggest that routing-based rejection predictors are unlikely to be useful in practice: routing inertia is statistically robust but offers limited leverage for early-exit decisions.

\section{Conclusion}
\label{sec:conclusion}
As sparse MoE models move into increasingly memory-constrained inference regimes, an important question is whether speculative decoding remains worthwhile at low batch size.
Across 10 recent sparse MoE models (30B--1T parameters, 8--512 experts), we find that the previously expected concave SD speedup curve does not characterize most modern sparse models: many attain high speedup at or near batch size~1.
We show that verification cost in sparse MoE models is not determined by activated-expert bytes alone, but is also shaped by temporal overlap across consecutive tokens and by GPU execution dynamics, whose specific mechanisms vary across hardware and kernels.
A maximum-entropy IPF framework predicts unique expert counts from routing-overlap statistics with less than 4\% error and, under synthetic routing with fixed execution conditions, substantially narrows the gap to real-model curves while capturing the qualitative low-batch trend, thereby providing a train-free way to study target and draft architecture tradeoffs.
Taken together, these results suggest that SD performance on sparse MoE models should be modeled using both routing-overlap statistics and GPU execution dynamics, and that fine-grained MoE designs may be more compatible with low-batch speculative inference than previously believed.

\begin{ack}
This work was funded by Cohere, where all authors were employed. We thank Zewen Shen for their helpful feedback on the paper.
\end{ack}

\bibliographystyle{plainnat}
\bibliography{references}

@article{leviathan2023fast,
  title={Fast inference from transformers via speculative decoding},
  author={Leviathan, Yaniv and Kalman, Matan and Matias, Yossi},
  journal={International Conference on Machine Learning},
  year={2023}
}

@article{chen2023accelerating,
  title={Accelerating large language model decoding with speculative sampling},
  author={Chen, Charlie and Borgeaud, Sebastian and Irving, Geoffrey and Lespiau, Jean-Baptiste and Sifre, Laurent and Jumper, John},
  journal={arXiv preprint arXiv:2302.01318},
  year={2023}
}

@article{li2024eagle,
  title={{EAGLE}: Speculative sampling requires rethinking feature uncertainty},
  author={Li, Yuhui and Wei, Fangyun and Zhang, Chao and Zhang, Hongyang},
  journal={International Conference on Machine Learning},
  year={2024}
}

@article{li2024eagle2,
  title={{EAGLE-2}: Faster inference of language models with dynamic draft trees},
  author={Li, Yuhui and Wei, Fangyun and Zhang, Chao and Zhang, Hongyang},
  journal={Proceedings of the 2024 Conference on Empirical Methods in Natural Language Processing},
  year={2024}
}

@article{li2025eagle3,
  title={{EAGLE-3}: Scaling up inference acceleration of large language models via training-time test},
  author={Li, Yuhui and Wei, Fangyun and Zhang, Chao and Zhang, Hongyang},
  journal={Advances in Neural Information Processing Systems},
  year={2025}
}

@article{cai2024medusa,
  title={Medusa: Simple {LLM} inference acceleration framework with multiple decoding heads},
  author={Cai, Tianle and Li, Yuhong and Geng, Zhengyang and Peng, Hongwu and Lee, Jason D and Chen, Deming and Dao, Tri},
  journal={International Conference on Machine Learning},
  year={2024}
}

@article{jiang2024mixtral,
  title={Mixtral of experts},
  author={Jiang, Albert Q and Sablayrolles, Alexandre and Roux, Antoine and Mensch, Arthur and Savary, Blanche and Bamford, Chris and Chaplot, Devendra Singh and Casas, Diego de las and Hanna, Emma Bou and Bressand, Florian and others},
  journal={arXiv preprint arXiv:2401.04088},
  year={2024}
}

@article{deepseekai2024deepseekv3,
  title={{DeepSeek-V3} technical report},
  author={{DeepSeek-AI}},
  journal={arXiv preprint arXiv:2412.19437},
  year={2024}
}

@article{yang2025qwen3,
  title={Qwen3 technical report},
  author={Yang, An and Yang, Baosong and Zhang, Beichen and others},
  journal={arXiv preprint arXiv:2505.09388},
  year={2025}
}

@article{huang2025moesd,
  title={{MoESD}: Unveil speculative decoding's potential for accelerating sparse {MoE}},
  author={Huang, Zongle and Zhu, Lei and Zhan, ZongYuan and Hu, Ting and Mao, Weikai and Yu, Xianzhi and Liu, Yongpan and Zhang, Tianyu},
  journal={Advances in Neural Information Processing Systems},
  year={2025}
}

@article{kwon2023vllm,
  title={Efficient memory management for large language model serving with {PagedAttention}},
  author={Kwon, Woosuk and Li, Zhuohan and Zhuang, Siyuan and Sheng, Ying and Zheng, Lianmin and Yu, Cody Hao and Gonzalez, Joseph E and Zhang, Hao and Stoica, Ion},
  journal={ACM SIGOPS Symposium on Operating Systems Principles},
  year={2023}
}

@article{kariyappa2026sidequest,
  title={{SideQuest}: Model-driven {KV} cache management for long-horizon agentic reasoning},
  author={Kariyappa, Sanjay and Suh, G. Edward},
  journal={arXiv preprint arXiv:2602.22603},
  year={2026}
}

@article{zheng2023mtbench,
  title={Judging {LLM}-as-a-Judge with {MT-Bench} and {Chatbot Arena}},
  author={Zheng, Lianmin and others},
  journal={Neural Information Processing Systems},
  year={2023}
}

@misc{vicuna2023,
  title={Vicuna: An Open-Source Chatbot Impressing {GPT-4} with 90\%* {ChatGPT} Quality},
  author={Chiang, Wei-Lin and Li, Zhuohan and Lin, Zi and Sheng, Ying and Wu, Zhanghao and
          Zhang, Hao and Zheng, Lianmin and Zhuang, Siyuan and Zhuang, Yonghao and
          Gonzalez, Joseph E. and Stoica, Ion and Xing, Eric P.},
  year={2023},
  month={March},
  url={https://lmsys.org/blog/2023-03-30-vicuna/}
}

@inproceedings{li2024instructcoder,
  title={InstructCoder: Instruction Tuning Large Language Models for Code Editing},
  author={Li, Kaixin and Hu, Qisheng and Zhao, James Xu and Chen, Hui and Xie, Yuxi and Liu, Tiedong and Shieh, Michael and He, Junxian},
  booktitle={Proceedings of the 62nd Annual Meeting of the Association for Computational Linguistics (Volume 4: Student Research Workshop)},
  pages={473--493},
  address={Bangkok, Thailand},
  publisher={Association for Computational Linguistics},
  year={2024},
  url={https://aclanthology.org/2024.acl-srw.52/}
}

@article{sadhukhan2024magicdec,
  title={{MagicDec}: Breaking the latency-throughput tradeoff for long context generation with speculative decoding},
  author={Sadhukhan, Ranajoy and Chen, Jian and Chen, Zhuoming and Tiwari, Vashisth and Lai, Ruihang and Shi, Jinyuan and Yen, Ian En-Hsu and May, Avner and Chen, Tianqi and Chen, Beidi},
  journal={International Conference on Learning Representations},
  year={2025}
}

@inproceedings{abramovich2026speedbench,
  title={{SPEED-Bench}: A Unified and Diverse Benchmark for Speculative Decoding},
  author={Abramovich, Talor and Ashkenazi, Maor and Putterman, Izzy and Chislett, Benjamin and Mitra, Tiyasa and Rouhani, Bita Darvish and Zilberstein, Ran and Geifman, Yonatan},
  booktitle={International Conference on Machine Learning},
  year={2026},
  note={arXiv:2604.09557}
}

@article{gloeckle2024multitoken,
  title={Better \& faster large language models via multi-token prediction},
  author={Gloeckle, Fabian and Idrissi, Badr Youbi and Rozi{\`e}re, Baptiste and Lopez-Paz, David and Synnaeve, Gabriel},
  journal={International Conference on Machine Learning},
  year={2024}
}

@article{olson2025probing,
  title={Probing semantic routing in large mixture-of-expert models},
  author={Olson, Matthew Lyle and Ratzlaff, Neale and Hinck, Musashi and Luo, Man and Yu, Sungduk and Xue, Chendi and Lal, Vasudev},
  journal={Findings of the Association for Computational Linguistics: EMNLP 2025},
  pages={18263--18278},
  year={2025}
}

@article{bershatsky2025spatial,
  title={On the spatial structure of mixture-of-experts in transformers},
  author={Bershatsky, Daniel and Oseledets, Ivan},
  journal={ICLR 2025 Workshop on Sparsity in {LLM}s},
  year={2025}
}

@article{he2024expertflow,
  title={{ExpertFlow}: Efficient mixture-of-experts inference via predictive expert caching and token scheduling},
  author={He, Xin and Zhang, Shunkang and Tang, Kaijie and Shi, Shaohuai and Wang, Yuxin and Zeng, Zihao and Tang, Zhenheng and Chu, Xiaowen and Yin, Haiyan and Tsang, Ivor W. and Ong, Yew Soon},
  journal={ACM/IEEE Design Automation Conference},
  year={2026}
}

@article{jiang2025scylla,
  title={Scaling laws for speculative decoding},
  author={Jiang, Guoqing and others},
  journal={arXiv preprint arXiv:2505.07858},
  year={2025}
}

@article{dao2022flashattention,
  title={{FlashAttention}: Fast and memory-efficient exact attention with {IO}-awareness},
  author={Dao, Tri and Fu, Daniel Y and Ermon, Stefano and Rudra, Atri and Re, Christopher},
  journal={Neural Information Processing Systems},
  year={2022}
}

@article{deming1940leastsquares,
  title={On a least squares adjustment of a sampled frequency table when the expected marginal totals are known},
  author={Deming, W. Edwards and Stephan, Frederick F.},
  journal={The Annals of Mathematical Statistics},
  year={1940}
}

@article{csiszar1975idivergence,
  title={$I$-Divergence geometry of probability distributions and minimization problems},
  author={Csiszar, I.},
  journal={The Annals of Probability},
  year={1975}
}

@article{xia2024specbench,
  title={Unlocking Efficiency in Large Language Model Inference: A Comprehensive Survey of Speculative Decoding},
  author={Xia, Heming and Yang, Zhe and Dong, Qingxiu and Wang, Peiyi and Li, Yongqi and Ge, Tao and Liu, Tianyu and Li, Wenjie and Sui, Zhifang},
  journal={Findings of the Association for Computational Linguistics: ACL 2024},
  year={2024},
  url={https://arxiv.org/abs/2401.07851}
}

@inproceedings{singh2024aya,
  title={Aya Dataset: An Open-Access Collection for Multilingual Instruction Tuning},
  author={Singh, Shivalika and Vargus, Freddie and D'Souza, Daniel and Karlsson, B{\"o}rje F. and Mahendiran, Abinaya and Ko, Wei-Yin and Shandilya, Herumb and Patel, Jay and Mataciunas, Deividas and O'Mahony, Laura and Zhang, Mike and Hettiarachchi, Ramith and Wilson, Joseph and Machado, Marina and Moura, Luisa and Krzemi{\'n}ski, Dominik and Fadaei, Hakimeh and Ergun, Irem and Okoh, Ifeoma and Alaagib, Aisha and Mudannayake, Oshan and Alyafeai, Zaid and Chien, Vu and Ruder, Sebastian and Guthikonda, Surya and Alghamdi, Emad and Gehrmann, Sebastian and Muennighoff, Niklas and Bartolo, Max and Kreutzer, Julia and {\"U}st{\"u}n, Ahmet and Fadaee, Marzieh and Hooker, Sara},
  booktitle={Proceedings of the 62nd Annual Meeting of the Association for Computational Linguistics (Volume 1: Long Papers)},
  year={2024},
  address={Bangkok, Thailand},
  publisher={Association for Computational Linguistics},
  url={https://aclanthology.org/2024.acl-long.620/},
  doi={10.18653/v1/2024.acl-long.620},
  pages={11521--11567}
}

@article{huang2026peagle,
  title={{P-EAGLE}: Parallel-drafting {EAGLE} with scalable training},
  author={Huang, Xin and others},
  journal={arXiv preprint arXiv:2602.01469},
  year={2026}
}

@article{tang2025llamasd,
  title={Efficient speculative decoding for {Llama} at scale: Challenges and solutions},
  author={Tang, Bangsheng and Fu, Carl Chengyan and Kou, Fei and Sizov, Grigory and Zhang, Haoci and Park, Jason and Liu, Jiawen and You, Jie and Yang, Qirui and Mehta, Sachin and others},
  journal={arXiv preprint arXiv:2508.08192},
  year={2025}
}

@misc{nvidia2025gptosseagle3,
  title={nvidia/gpt-oss-120b-Eagle3-long-context},
  author={{NVIDIA}},
  year={2025},
  url={https://huggingface.co/nvidia/gpt-oss-120b-Eagle3-long-context},
  note={Hugging Face model card}
}

@misc{redhat2026gptossspec,
  title={Performance improvements with speculative decoding in vLLM for gpt-oss},
  author={{Red Hat Developer}},
  year={2026},
  url={https://developers.redhat.com/articles/2026/04/16/performance-improvements-speculative-decoding-vllm-gpt-oss#get_started},
  note={Technical article}
}

@article{kimik25,
  title={Kimi {K2.5}: Visual agentic intelligence},
  author={Du, Yulun and others},
  journal={arXiv preprint arXiv:2602.02276},
  year={2026}
}

@article{xia2023specdec,
  title={Speculative decoding: Exploiting speculative execution for accelerating seq2seq generation},
  author={Xia, Heming and Ge, Tao and Wang, Peiyi and Chen, Si-Qing and Wei, Furu and Sui, Zhifang},
  journal={Findings of the Association for Computational Linguistics: EMNLP 2023},
  year={2023}
}

@article{stern2018blockwise,
  title={Blockwise parallel decoding for deep autoregressive models},
  author={Stern, Mitchell and Shazeer, Noam and Uszkoreit, Jakob},
  journal={Neural Information Processing Systems},
  year={2018}
}

@article{sun2023spectr,
  title={{SpecTr}: Fast speculative decoding via optimal transport},
  author={Sun, Ziteng and Suresh, Ananda Theertha and Ro, Jae Hun and Beirami, Ahmad and Jain, Himanshu and Yu, Felix},
  journal={Neural Information Processing Systems},
  year={2023}
}

@article{fu2024lookahead,
  title={Break the sequential dependency of {LLM} inference using lookahead decoding},
  author={Fu, Yichao and Bailis, Peter and Stoica, Ion and Zhang, Hao},
  journal={International Conference on Machine Learning},
  year={2024}
}

@article{ankner2024hydra,
  title={{Hydra}: Sequentially-dependent draft heads for Medusa decoding},
  author={Ankner, Zachary and Parthasarathy, Rishab and Nrusimha, Aniruddha and Rinard, Christopher and Ragan-Kelley, Jonathan and Brandon, William},
  journal={Conference on Language Modeling},
  year={2024}
}

@article{miao2024specinfer,
  title={{SpecInfer}: Accelerating large language model serving with tree-based speculative inference and verification},
  author={Miao, Xupeng and Oliaro, Gabriele and Zhang, Zhihao and Cheng, Xinhao and Wang, Zeyu and Wong, Rae Ying Yee and Chen, Zhuoming and Arfeen, Daiyaan and Abhyankar, Reyna and Jia, Zhihao},
  journal={ACM International Conference on Architectural Support for Programming Languages and Operating Systems},
  year={2024}
}

@article{he2024rest,
  title={{REST}: Retrieval-based speculative decoding},
  author={He, Zhenyu and Zhong, Zexuan and Cai, Tianle and Lee, Jason and He, Di},
  journal={North American Chapter of the Association for Computational Linguistics: Human Language Technologies},
  year={2024}
}

@article{svirschevski2024specexec,
  title={{SpecExec}: Massively parallel speculative decoding for interactive {LLM} inference on consumer devices},
  author={Svirschevski, Ruslan and May, Avner and Chen, Zhuoming and Chen, Beidi and Jia, Zhihao and Ryabinin, Max},
  journal={Neural Information Processing Systems},
  year={2024}
}

@article{shazeer2017outrageously,
  title={Outrageously large neural networks: The sparsely-gated mixture-of-experts layer},
  author={Shazeer, Noam and Mirhoseini, Azalia and Maziarz, Krzysztof and Davis, Andy and Le, Quoc V and Hinton, Geoffrey and Dean, Jeff},
  journal={International Conference on Learning Representations},
  year={2017}
}

@article{lepikhin2021gshard,
  title={{GShard}: Scaling giant models with conditional computation and automatic sharding},
  author={Lepikhin, Dmitry and Lee, HyoukJoong and Xu, Yuanzhong and Chen, Dehao and Firat, Orhan and Huang, Yanping and Krikun, Maxim and Shazeer, Noam and Chen, Zhifeng},
  journal={International Conference on Learning Representations},
  year={2021}
}

@article{lewis2021base,
  title={{BASE} layers: Simplifying training of large, sparse models},
  author={Lewis, Mike and Bhosale, Shruti and Dettmers, Tim and Goyal, Naman and Zettlemoyer, Luke},
  journal={International Conference on Machine Learning},
  year={2021}
}

@article{du2022glam,
  title={{GLaM}: Efficient scaling of language models with mixture-of-experts},
  author={Du, Nan and others},
  journal={International Conference on Machine Learning},
  year={2022}
}

@article{fedus2022switch,
  title={Switch transformers: Scaling to trillion parameter models with simple and efficient sparsity},
  author={Fedus, William and Zoph, Barret and Shazeer, Noam},
  journal={Journal of Machine Learning Research},
  year={2022}
}

@article{zhou2022expert,
  title={Mixture-of-experts with expert choice routing},
  author={Zhou, Yanqi and Lei, Tao and Liu, Hanxiao and Du, Nan and Huang, Yanping and Zhao, Vincent and Dai, Andrew M and Chen, Zhifeng and Le, Quoc V and Laudon, James},
  journal={Neural Information Processing Systems},
  year={2022}
}

@article{komatsuzaki2023upcycling,
  title={Sparse upcycling: Training mixture-of-experts from dense checkpoints},
  author={Komatsuzaki, Aran and Puigcerver, Joan and Lee-Thorp, James and Riquelme, Carlos and Mustafa, Basil and Ainslie, Joshua and Tay, Yi and Dehghani, Mostafa and Houlsby, Neil},
  journal={International Conference on Learning Representations},
  year={2023}
}

@inproceedings{dai2024deepseekmoe,
  title={{DeepSeekMoE}: Towards Ultimate Expert Specialization in Mixture-of-Experts Language Models},
  author={Dai, Damai and Deng, Chengqi and Zhao, Chenggang and Xu, R.X. and Gao, Huazuo and Chen, Deli and Li, Jiashi and Zeng, Wangding and Yu, Xingkai and Wu, Y. and Xie, Zhenda and Li, Y.K. and Huang, Panpan and Luo, Fuli and Ruan, Chong and Sui, Zhifang and Liang, Wenfeng},
  booktitle={Proceedings of the 62nd Annual Meeting of the Association for Computational Linguistics (Volume 1: Long Papers)},
  pages={1280--1297},
  address={Bangkok, Thailand},
  publisher={Association for Computational Linguistics},
  doi={10.18653/v1/2024.acl-long.70},
  url={https://aclanthology.org/2024.acl-long.70/},
  year={2024}
}

@article{dao2024flashattention2,
  title={{FlashAttention-2}: Faster attention with better parallelism and work partitioning},
  author={Dao, Tri},
  journal={International Conference on Learning Representations},
  year={2024}
}

@article{williams2009roofline,
  title={Roofline: An insightful visual performance model for multicore architectures},
  author={Williams, Samuel and Waterman, Andrew and Patterson, David},
  journal={Communications of the ACM},
  year={2009}
}

@inproceedings{ludziejewski2024scaling,
  title={Scaling Laws for Fine-Grained Mixture of Experts},
  author={Ludziejewski, Jan and Krajewski, Jakub and Adamczewski, Kamil and Pi\'{o}ro, Maciej and Krutul, Micha{\l} and Antoniak, Szymon and Ciebiera, Kamil and Kr\'{o}l, Krystian and Odrzyg\'{o}\'{z}d\'{z}, Tomasz and Sankowski, Piotr and Cygan, Marek and Jaszczur, Sebastian},
  booktitle={Proceedings of the 41st International Conference on Machine Learning},
  series={Proceedings of Machine Learning Research},
  volume={235},
  pages={33270--33288},
  publisher={PMLR},
  url={https://proceedings.mlr.press/v235/ludziejewski24a.html},
  year={2024}
}

@article{deepseek2025r1,
  title={{DeepSeek-R1} incentivizes reasoning in {LLM}s through reinforcement learning},
  author={{DeepSeek-AI}},
  journal={Nature},
  year={2025}
}

@article{kimi2025k15,
  title={Kimi k1.5: Scaling reinforcement learning with {LLM}s},
  author={{Kimi Team}},
  journal={arXiv preprint arXiv:2501.12599},
  year={2025}
}

@inproceedings{snell2025scaling,
  title={Scaling {LLM} test-time compute optimally can be more effective than scaling parameters for reasoning},
  author={Snell, Charlie and Lee, Jaehoon and Xu, Kelvin and Kumar, Aviral},
  booktitle={International Conference on Learning Representations},
  year={2025}
}

@article{ulorl2025,
  title={{UloRL}: An ultra-long output reinforcement learning approach for advancing large language models' reasoning abilities},
  author={Du, Dong and Liu, Shulin and Yang, Tao and Chen, Shaohua and Li, Yang},
  journal={arXiv preprint arXiv:2507.19766},
  year={2025}
}

@article{sun2025contextfolding,
  title={Scaling long-horizon {LLM} agent via context-folding},
  author={Sun, Weiwei and Lu, Miao and Ling, Zhan and Liu, Kang and Yao, Xuesong and Yang, Yiming and Chen, Jiecao},
  journal={arXiv preprint arXiv:2510.11967},
  year={2025}
}

@article{saxena2025utility,
  title={Utility-Driven Speculative Decoding for Mixture-of-Experts},
  author={Saxena, Anish and Tsai, Po-An and Taneja, Hritvik and Jaleel, Aamer and Qureshi, Moinuddin},
  journal={IEEE/ACM International Symposium on Microarchitecture},
  year={2025}
}

\appendix

\section{Limitations}
\label{sec:limitations}

Our synthetic-routing analysis is approximate.
The IPF construction matches pairwise overlap statistics, but does not model higher-order routing structure, position dependence in expert marginals, or how temporal overlap itself varies with architectural choices.
In practice, one may use overlap statistics from a closely related model family, or aggregate statistics from the models we study, as a better train-free estimate than a uniform baseline when exploring target and draft design tradeoffs before training.
However, understanding which architectural factors determine temporal overlap, and how reliably overlap transfers across model families and sparsity patterns, remains an important direction for future work.

\section{Experimental Setup}
\label{app:reproducibility}

All reported experiments use vLLM v1 (commit \texttt{7c080dd3c}) with $K{=}3$ draft tokens (except the $K$-sweep in Section~\ref{sec:k-sweep}, which uses $K\in\{3,5,7\}$) and MT-Bench as the base dataset.
For the routing analyses (Section~\ref{sec:temporal} and Appendix~\ref{app:routing-inertia}), we build on this commit and extend its \texttt{enable\_return\_routed\_experts} routing capture to also record target-model routing at rejected draft positions after each verification step.
For the end-to-end SD sweep, the benchmark client oversamples the 80 MT-Bench prompts with replacement to match the configured number of requests at each concurrency setting, uses fixed 256-token outputs, and runs with prefix caching disabled.
Every end-to-end speedup measurement was repeated three times in independent server sessions, and the coefficient of variation (CV, std/mean) of the speedup is below 5\% at every batch size for every configuration (e.g., at most 1.4\% for Mixtral-8x7B FP8 TP1 and Qwen3-30B FP8 TP2, and at most 3.3\% for Qwen3-235B FP8 TP4).
Because the acceptance length is nearly constant across runs, target efficiency (speedup divided by acceptance length) inherits the same low variability; figures and tables report a single run.
Table~\ref{tab:models} summarizes the evaluated MoE models and their SD methods, and Table~\ref{tab:bench-schedule} summarizes the benchmark schedule.
All models run on NVIDIA H100 SXM GPUs except for Kimi-K2.5 and DeepSeek-V3 which use NVIDIA B200 GPUs.
Tables~\ref{tab:model-target-weights} and~\ref{tab:model-draft-weights} list the Hugging Face model identifiers, draft weights, SD method, tensor parallelism, and GPU used for each reported configuration.
All quantized configurations (FP8, INT4, and MXFP4) use the quantization recipe stored in the corresponding published Hugging Face checkpoint, loaded as-is by vLLM; we apply no additional quantization.

\begin{table}[h]
  \caption{MoE models evaluated.
  Here $N$ denotes the total number of experts, $k$ the number of experts per token, $E_\text{dim}$ the expert intermediate size, $\text{SE}_\text{dim}$ the shared-expert intermediate size (\texttt{n/a} if none), and TP the tensor parallelism.}
  \label{tab:models}
  \centering
  \small
  \begin{tabular}{lcccccccl}
    \toprule
    Model & $k/N$ & $E_\text{dim}$ & $\text{SE}_\text{dim}$ & Hidden & Layers & TP & SD Method \\
    \midrule
    Mixtral-8x7B       & 2/8    & 14336 & n/a   & 4096 & 32 & 2 & EAGLE \\
    Qwen2-57B          & 8/64   & 2560  & 20480 & 3584 & 28 & 2 & Draft model (0.5B) \\
    GLM-4.7-Flash      & 4/64   & 1536  & 10240 & 2048 & 40 & 1 & MTP \\
    Qwen3-30B          & 8/128  & 768   & n/a   & 2048 & 48 & 1 & EAGLE-3 \\
    Qwen3-235B         & 8/128  & 1536  & n/a   & 4096 & 94 & 4 & EAGLE-3 \\
    GPT-OSS-120B       & 4/128  & 2880  & n/a   & 2880 & 56 & 4 & EAGLE-3 \\
    DeepSeek-V3        & 8/256  & 2048  & 18432 & 7168 & 61 & 8 & MTP \\
    Kimi-K2.5          & 8/384  & 2048  & 18432 & 7168 & 61 & 8 & EAGLE-3 \\
    Qwen3.5-35B        & 8/256  & 512   & 512   & 2048 & 64 & 2 & MTP \\
    Qwen3-Next-80B     & 10/512 & 512   & 512   & 2048 & 80 & 4 & MTP \\
    \bottomrule
  \end{tabular}
\end{table}

\begin{table}[h]
  \caption{Benchmark schedule for the MT-Bench end-to-end SD sweep.
  The 80 base prompts are oversampled with replacement when the configured number of requests exceeds 80; all runs use fixed 256-token outputs and disable prefix caching.}
  \label{tab:bench-schedule}
  \centering
  \scriptsize
  \begin{tabular}{rrrr}
    \toprule
    Batch size & Num prompts & Batch size & Num prompts \\
    \midrule
    1 & 80   & 16  & 320  \\
    2 & 80   & 32  & 640  \\
    4 & 80   & 64  & 1280 \\
    8 & 160  & 128 & 2560 \\
    \bottomrule
  \end{tabular}
\end{table}

\begin{table}[h]
  \caption{Target-weight identifiers and runtime configuration used in the reported experiments.}
  \label{tab:model-target-weights}
  \centering
  \scriptsize
  \begin{tabular}{llcc}
    \toprule
    Model & Target weights & TP & GPU \\
    \midrule
    Mixtral-8x7B
      & \texttt{mistralai/Mixtral-8x7B-Instruct-v0.1}
      & 2 & H100 \\
    Qwen2-57B
      & \texttt{Qwen/Qwen2-57B-A14B-Instruct}
      & 2 & H100 \\
    GLM-4.7-Flash
      & \texttt{zai-org/GLM-4.7-Flash}
      & 1 & H100 \\
    Qwen3-30B
      & \texttt{Qwen/Qwen3-30B-A3B-FP8}
      & 1 & H100 \\
    Qwen3-235B
      & \texttt{Qwen/Qwen3-235B-A22B-Instruct-2507-FP8}
      & 4 & H100 \\
    GPT-OSS-120B
      & \texttt{openai/gpt-oss-120b}
      & 4 & H100 \\
    DeepSeek-V3
      & \texttt{deepseek-ai/DeepSeek-V3-0324}
      & 8 & B200 \\
    Kimi-K2.5
      & \texttt{moonshotai/Kimi-K2.5}
      & 8 & B200 \\
    Qwen3.5-35B
      & \texttt{Qwen/Qwen3.5-35B-A3B}
      & 2 & H100 \\
    Qwen3-Next-80B
      & \texttt{Qwen/Qwen3-Next-80B-A3B-Instruct}
      & 4 & H100 \\
    \bottomrule
  \end{tabular}
\end{table}

\begin{table}[h]
  \caption{Draft-weight identifiers and SD methods used in the reported experiments.}
  \label{tab:model-draft-weights}
  \centering
  \scriptsize
  \begin{tabular}{lll}
    \toprule
    Model & Draft weights & Method \\
    \midrule
    Mixtral-8x7B
      & \texttt{yuhuili/EAGLE-mixtral-instruct-8x7B}
      & EAGLE \\
    Qwen2-57B
      & \texttt{Qwen/Qwen2-0.5B-Instruct}
      & Draft model \\
    GLM-4.7-Flash
      & (built-in MTP heads)
      & MTP \\
    Qwen3-30B
      & \texttt{RedHatAI/Qwen3-30B-A3B-Instruct-2507-speculator.eagle3}
      & EAGLE-3 \\
    Qwen3-235B
      & \texttt{RedHatAI/Qwen3-235B-A22B-Instruct-2507-speculator.eagle3}
      & EAGLE-3 \\
    GPT-OSS-120B
      & \texttt{nvidia/gpt-oss-120b-Eagle3-long-context}
      & EAGLE-3 \\
    DeepSeek-V3
      & (built-in MTP heads)
      & MTP \\
    Kimi-K2.5
      & \texttt{AQ-MedAI/Kimi-K25-eagle3}
      & EAGLE-3 \\
    Qwen3.5-35B
      & (built-in MTP heads)
      & MTP \\
    Qwen3-Next-80B
      & (built-in MTP heads)
      & MTP \\
    \bottomrule
  \end{tabular}
\end{table}

\section{Consolidated Target Efficiency Measurements}
\label{app:rfwd}

\begin{table}[h]
  \caption{Target efficiency at batch size 1 for $K{=}3$.
  Higher values indicate non-concave speedup curves as observed in Figure~\ref{fig:mega-grid}.}
  \label{tab:target-efficiency-appendix}
  \centering
  \scriptsize
  \setlength{\tabcolsep}{3pt}
  \resizebox{\linewidth}{!}{%
  \begin{tabular}{llc@{\hspace{8mm}}llc}
    \toprule
    Model & SD method & Target eff. & Model & SD method & Target eff. \\
    \midrule
    Qwen2-57B (bf16, TP2)      & Separate draft & 0.270 & Qwen3-30B (fp8, TP2)      & EAGLE-3 & 0.860 \\
    Mixtral-8x7B (bf16, TP2)   & Separate draft & 0.192 & Qwen3-235B (bf16, TP8)    & EAGLE-3 & 0.786 \\
    Mixtral-8x7B (bf16, TP2)   & EAGLE          & 0.505 & Qwen3-235B (fp8, TP4)     & EAGLE-3 & 0.889 \\
    Mixtral-8x7B (fp8, TP1)    & EAGLE          & 0.488 & Qwen3-Next-80B (bf16, TP4)& MTP     & 0.615 \\
    GLM-4.7-Flash (bf16, TP1)  & MTP            & 0.554 & Qwen3-Next-80B (fp8, TP2) & MTP     & 0.707 \\
    GLM-4.7-Flash (bf16, TP2)  & MTP            & 0.568 & Qwen3.5-35B (bf16, TP2)   & MTP     & 0.546 \\
    Qwen3-30B (bf16, TP1)      & EAGLE-3        & 0.630 & Kimi-K2.5 (int4, TP8)     & EAGLE-3 & 0.752 \\
    Qwen3-30B (bf16, TP2)      & EAGLE-3        & 0.697 & DeepSeek-V3 (fp8, TP8)    & MTP     & 0.785 \\
    Qwen3-30B (fp8, TP1)       & EAGLE-3        & 0.818 & GPT-OSS-120B (mxfp4, TP4) & EAGLE-3 & 0.810 \\
    \bottomrule
  \end{tabular}%
  }
\end{table}

\begin{figure}[h]
  \centering
  \includegraphics[width=\linewidth]{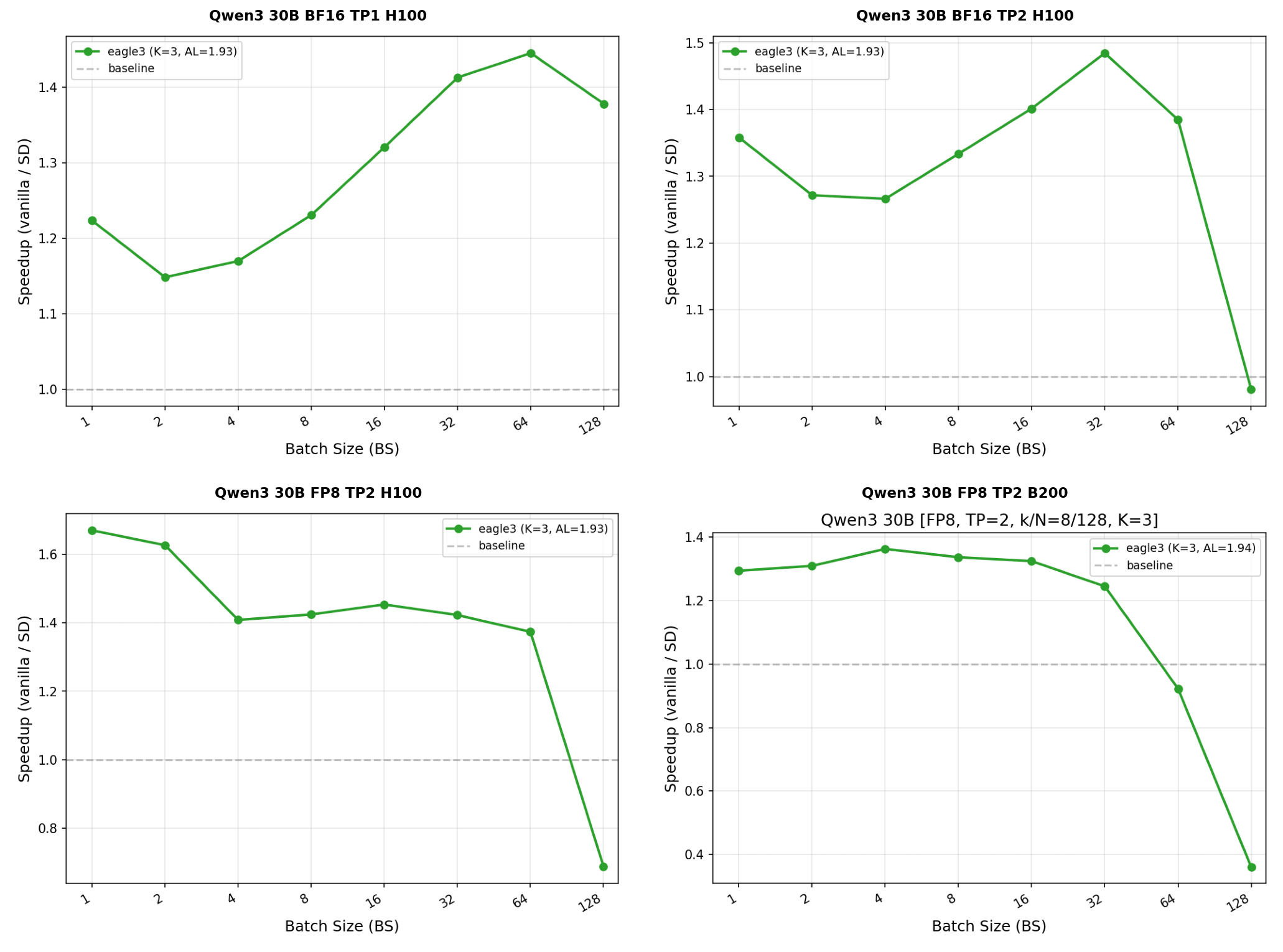}
  \caption{Qwen3-30B TPOT speedup versus batch size across TP/precision settings and hardware.
  Panels show BF16 TP1 H100, BF16 TP2 H100, FP8 TP2 H100, and FP8 TP2 B200 (126\,MB L2).}
  \label{fig:qwen3-30b-config-sweep}
\end{figure}

\paragraph{Qwen3-30B configuration and hardware sweep.}
Figure~\ref{fig:qwen3-30b-config-sweep} compares Qwen3-30B speedup curves across BF16/FP8 and TP1/TP2 settings on H100, together with an FP8, TP2 run on B200.
The comparison complements Table~\ref{tab:target-efficiency-appendix} by showing how the same model family's batch-size trajectory changes as per-GPU expert shards shrink and as the hardware memory hierarchy changes.

\subsection{Full 10-Model Speedup Grid}

\begin{figure}[ht]
  \centering
  \includegraphics[width=\textwidth]{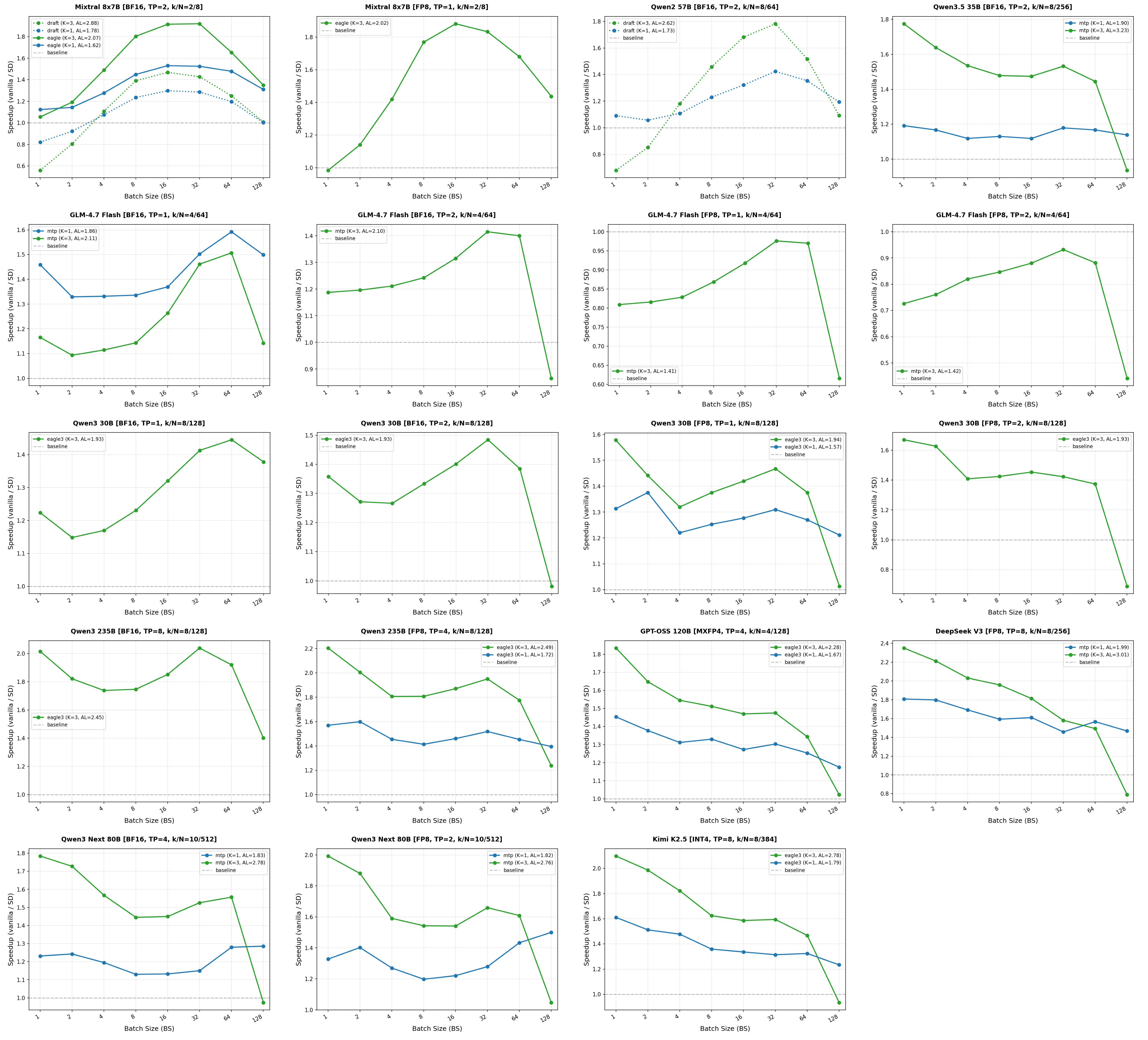}
  \caption{TPOT speedup versus batch size for all 10 MoE models ($K{=}3$).}
  \label{fig:mega-grid}
\end{figure}

Figure~\ref{fig:mega-grid} plots TPOT speedup versus batch size for all 10 models, complementing the four representative examples in Figure~\ref{fig:speedup-curves}.

%
%
%

\section{Cross-Dataset Stability of Non-Concave Speedup Curves}
\label{app:multi-dataset-speedup}

\begin{figure}[ht]
  \centering
  \captionsetup[subfigure]{font=footnotesize, justification=centering}
  \begin{subfigure}[t]{0.48\textwidth}
    \centering
    \includegraphics[width=\linewidth]{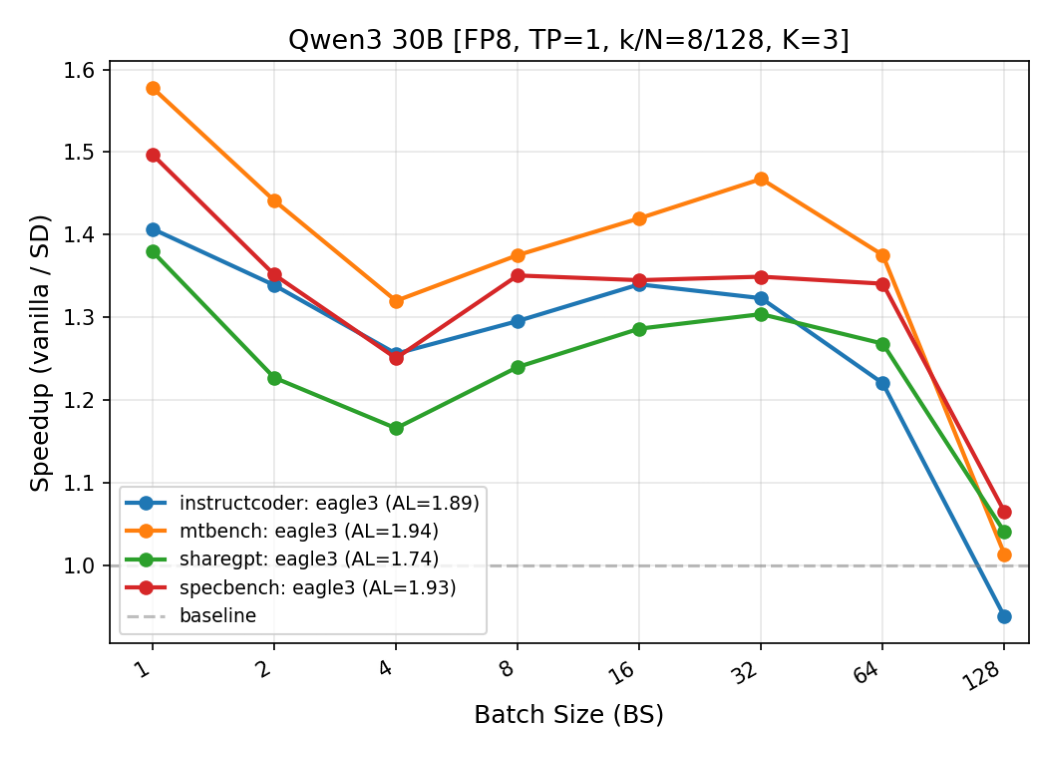}
    \caption{Qwen3-30B}
  \end{subfigure}
  \hfill
  \begin{subfigure}[t]{0.48\textwidth}
    \centering
    \includegraphics[width=\linewidth]{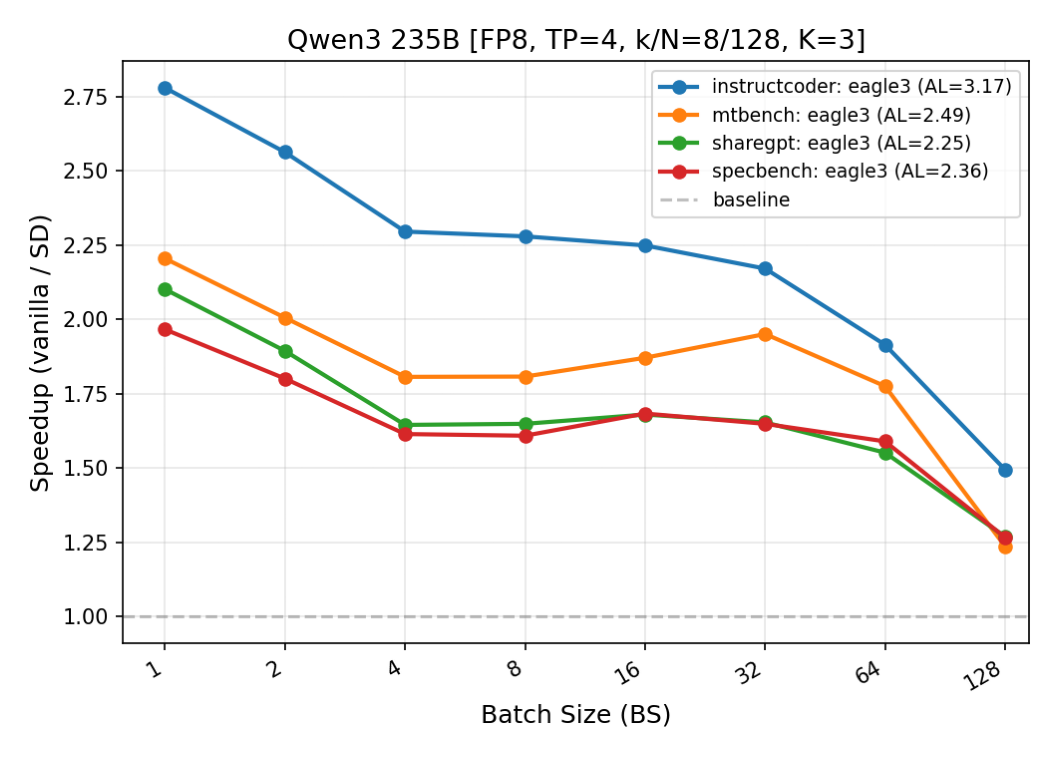}
    \caption{Qwen3-235B}
  \end{subfigure}
  \caption{TPOT speedup versus batch size across four datasets for two representative non-concave models.
  Each curve corresponds to InstructCoder~\citep{li2024instructcoder}, MT-Bench~\citep{zheng2023mtbench}, ShareGPT~\citep{vicuna2023}, or Spec-Bench~\citep{xia2024specbench}. Although the absolute speedups vary by dataset, both models retain the same qualitative non-concave, batch-size-1-peaked behavior.}
  \label{fig:multi-dataset-speedup}
\end{figure}

Figure~\ref{fig:multi-dataset-speedup} shows that the non-concave, batch-size-1-peaked speedup pattern persists across InstructCoder~\citep{li2024instructcoder}, MT-Bench~\citep{zheng2023mtbench}, ShareGPT~\citep{vicuna2023}, and Spec-Bench~\citep{xia2024specbench} for Qwen3-30B and Qwen3-235B.
This robustness indicates that the low-batch speedup advantage is not an artifact of the MT-Bench prompt distribution used in the main paper.
In turn, it supports the interpretation that the favorable low-batch behavior of recent fine-grained MoEs reflects stable target-side execution properties rather than a dataset-specific acceptance effect.
Although the absolute speedups vary by dataset, the qualitative curve shape remains the same, so the main conclusion is about regime structure rather than any single benchmark's exact TPOT values.

\section{Three-Step Correction: From Weight Model to Measured Target Efficiency}
\label{app:three-step}

For Qwen3-30B at batch size~1 on H100, the activated-expert-bytes model predicts $R_\text{fwd}{=}0.51$, whereas the measured value is 0.87--0.90.
The gap decomposes into three independently measured corrections using the identity
\begin{equation}
  R_\text{fwd} = \frac{1}{f_\text{moe} \cdot S_\text{moe} + f_\text{non\text{-}moe} \cdot S_\text{non\text{-}moe}},
  \label{eq:rfwd-decomp}
\end{equation}
where $f$ denotes the fraction of vanilla forward GPU time and $S$ denotes the SD scaling factor for each component.

\paragraph{Step 0: Activated-expert-bytes baseline ($R{=}0.511$).}
This baseline uses weight bytes as a cost proxy: $f_\text{moe}{=}66.7\%$, $S_\text{moe}{=}2.44$ (linear in the unique expert count), and $S_\text{non\text{-}moe}{=}1.0$.
$$R = 1/(0.667 \times 2.44 + 0.333 \times 1.0) = 1/1.96 = 0.511$$

\paragraph{Step 1: Correcting the MoE fraction ($R$: 0.51 $\to$ 0.68, $+$0.168, accounting for 44\% of the gap).}
Kernel-level profiling measures the MoE pipeline at $1629/4957 = 32.9\%$ of vanilla target forward GPU time, rather than the 66.7\% predicted by weight bytes (\S\ref{sec:gpu-execution-dynamics}).
$$R = 1/(0.329 \times 2.44 + 0.671 \times 1.0) = 1/1.47 = 0.679$$

\paragraph{Step 2: Correcting the MoE scaling ($R$: 0.68 $\to$ 0.84, $+$0.164, accounting for 43\% of the gap).}
Kernel-level profiling measures the MoE pipeline scaling at $S{=}1.57$ rather than the weight-linear value of $S{=}2.44$, due to L2 cache reuse (5\% $\to$ 31\% hit rate) and dilution of pipeline overhead (routing $S{=}1.0$, activation $S{=}1.13$).
$$R = 1/(0.329 \times 1.57 + 0.671 \times 1.0) = 1/1.19 = 0.843$$

\paragraph{Step 3: Correcting non-MoE scaling ($R$: 0.84 $\to$ 0.87, $+$0.029, accounting for 8\% of the gap).}
Non-MoE target components are slightly cheaper under SD: attention GEMMs have $S{=}1.01$, FlashAttention~\citep{dao2022flashattention} has $S{=}1.03$, but framework overhead is amortized ($S{=}0.67$), yielding a collective $S{=}0.94$.
$$R = 1/(0.329 \times 1.57 + 0.671 \times 0.94) = 1/1.15 = 0.872$$

Table~\ref{tab:three-step} summarizes the correction chain.
Each step uses independently measured data from different profiling methods.
The remaining gap from 0.872 (kernel-level sum) to 0.896 (sweep wall-clock) arises from inter-kernel scheduling gaps within the CUDA graph.

\begin{table}[h]
  \caption{Three-step correction for Qwen3-30B at batch size~1 on H100.}
  \label{tab:three-step}
  \centering
  \small
  \begin{tabular}{clccc}
    \toprule
    Step & Correction & Target efficiency & $\Delta R$ & \% gap closed \\
    \midrule
    0 & Weight model baseline                                    & 0.511 & n/a    & n/a \\
    1 & MoE fraction: 67\% $\to$ 33\% (GPU time $\neq$ weight)  & 0.679 & +0.168 & 44\% \\
    2 & MoE scaling: $S$=2.44 $\to$ 1.57 (L2 cache + dilution)  & 0.843 & +0.164 & 43\% \\
    3 & Non-MoE: $S$=1.0 $\to$ 0.94 (amortization)              & 0.872 & +0.029 & 8\% \\
    \midrule
      & Measured (kernel-level sum)                             & \textbf{0.872} & & \\
      & Measured (sweep wall-clock)                              & \textbf{0.896} & & \\
    \bottomrule
  \end{tabular}
\end{table}

\section{Additional GPU Profiling: Qwen3-235B and B200}
\label{app:profiling-extra}

Table~\ref{tab:profiling-extra} extends the H100 profiling of Table~\ref{tab:profiling} to a second fine-grained model on H100 (Qwen3-235B, FP8, TP4) and to Qwen3-30B and Mixtral on B200 (FP8, TP1), using the same Nsight Systems and Nsight Compute methodology at batch size~1.
Each configuration uses the MoE kernel that vLLM selects by default.
The main result is that \emph{the activated-expert-bytes model fails as a cost model for sparse MoE verification in every configuration we profile}: it underpredicts target efficiency by 0.18--0.38 across all three configurations in Table~\ref{tab:profiling-extra}, in addition to the H100 gaps of Table~\ref{tab:profiling}, and the error is largest for fine-grained models with small experts.
The mechanisms behind the error differ across hardware and kernels, but the error itself persists.
Qwen3-235B reproduces both H100 mechanisms: MoE kernels achieve higher bandwidth than attention kernels, and the L2 hit rate rises under SD.

\paragraph{B200.}
On B200, the phenomenon reproduces: MoE accounts for 35\% and 69\% of GPU time for Qwen3-30B and Mixtral (vs.\ 33\% and 75\% on H100), and the activated-expert-bytes model still underpredicts target efficiency (0.76 and 0.61 observed vs.\ 0.51 and 0.43 predicted), with the larger gap again for the smaller experts.
The H100 mechanisms, however, do not carry over.
The default MoE kernel for Qwen3-30B on B200 groups co-routed tokens by expert and reads each expert's weights once, so there are no repeated reads for L2 to absorb and the L2 hit rate does not rise under SD (2.49\%$\to$2.21\%); attention and MoE kernels both run at only 12--15\% of peak bandwidth, leaving no attention-vs.-MoE disparity.
The L2 hit rate thus reflects kernel design rather than hardware: running the default H100 MoE kernel on B200 with the same IPF routing gives an SD L2 hit rate of 27.19\% for Qwen3-30B FP8, close to the 27.79\% measured on H100.
This suggests that at batch size~1 on B200, step time is dominated by costs that do not scale with expert bytes, likely fixed per-step overheads.
The common principle is that runtime depends on the bandwidth each kernel achieves and on fixed per-step costs, whereas the activated-expert-bytes model assumes weight bytes at a single constant bandwidth: on H100 achieved bandwidth differs sharply between kernels, and on B200 fixed costs dominate.
Verification cost should therefore be measured on the target hardware rather than derived from weight bytes, which is what the synthetic-routing method in Section~\ref{sec:end-to-end-validation-via-synthetic-routing} does.

\begin{table}[h]
  \caption{GPU profiling at batch size~1 for Qwen3-235B on H100 and for Qwen3-30B and Mixtral on B200 (all FP8). H100 has 132 SMs and B200 has 148 SMs.}
  \label{tab:profiling-extra}
  \centering
  \footnotesize
  \setlength{\tabcolsep}{4pt}
  \begin{tabular}{lccc}
    \toprule
    Metric & Qwen3-235B (H100, TP4) & Qwen3-30B (B200, TP1) & Mixtral-8x7B (B200, TP1) \\
    \midrule
    MoE weight reads (SD)    & per (token, expert) pair & once per expert & once per expert \\
    Expert size (per GPU)    & 4.5\,MB           & 4.5\,MB           & 168\,MB \\
    MoE share of weight      & 67.9\%            & 66.7\%            & 89.4\% \\
    MoE share of GPU time    & \textbf{34\%}     & \textbf{35\%}     & 69\% \\[3pt]
    Attn CTAs / SMs          & 18~(14\%)         & 16~(11\%)         & 48~(32\%) \\
    Attn HBM BW (\% peak)    & 19--26\%          & 12--13\%          & 27--37\% \\
    MoE CTAs / SMs           & 48--256           & 96--128           & 148 \\
    MoE HBM BW (\% peak)     & 34--46\%          & ${\sim}$15\%      & ${\sim}$23\% \\[3pt]
    Attn SD scaling $S$      & 1.00              & 0.97              & 0.98 \\
    MoE SD scaling $S$       & 1.51              & 1.90              & 1.91 \\
    L2 hit rate (AR$\to$SD)  & 5.70\%$\to$\textbf{28.66\%} & 2.49\%$\to$2.21\% & 14.78\%$\to$14.75\% \\[3pt]
    TE pred.                 & 0.51              & 0.51              & 0.43 \\
    TE observed              & 0.89              & 0.76              & 0.61 \\[3pt]
    Weight model gap         & +0.38             & +0.25             & +0.18 \\
    \bottomrule
  \end{tabular}
\end{table}

\section{Three-Level Overlap Decomposition Formulas}
\label{app:overlap-formulas}

We decompose expert overlap into three levels with increasing realism.
Each token selects exactly $k$ experts from $N$ total experts per layer; $p_e$ denotes the empirical selection probability of expert $e$ ($\sum_e p_e = k$).

\paragraph{Level 1: Uniform baseline.}
Assumes $p_e = k/N$ for all experts and that tokens are routed independently.
\begin{align}
  \text{overlap}_\text{uniform} &= \frac{k}{N}, \label{eq:uniform-overlap} \\
  \text{unique}_\text{uniform}(B) &= N \left[1 - \left(1 - \frac{k}{N}\right)^{\!B}\right].
  \label{eq:uniform-unique}
\end{align}

\paragraph{Level 2: Independence baseline.}
Uses the empirical per-layer expert probabilities $\{p_e\}$ while assuming that tokens are routed independently across timesteps.
This captures overlap from non-uniform expert popularity.
\begin{align}
  \text{overlap}_\text{indep} &= \frac{1}{k} \sum_e p_e^{\,2}, \label{eq:indep-overlap} \\
  \text{unique}_\text{indep}(B) &= \sum_e \left[1 - (1 - p_e)^B\right].
  \label{eq:indep-unique}
\end{align}

\paragraph{Level 3: Empirical.}
Measured directly from consecutive tokens while preserving temporal overlap:
\begin{align}
  \text{overlap}_\text{emp}(j) &= \frac{1}{k}\,\mathbb{E}\bigl[|\text{experts}(i) \cap \text{experts}(i{+}j)|\bigr], \label{eq:emp-overlap} \\
  \text{unique}_\text{emp}(B) &= \mathbb{E}\!\left[\left|\bigcup_{t=1}^{B} \text{experts}(t)\right|\right].
  \label{eq:emp-unique}
\end{align}

\paragraph{Decomposition.}
The empirical overlap can be decomposed as:
\begin{equation}
  \underbrace{\text{overlap}_\text{emp}}_{\text{observed}}
  = \underbrace{\frac{k}{N}}_{\text{combinatorial}}
  + \underbrace{\left(\text{overlap}_\text{indep} - \frac{k}{N}\right)}_{\text{expert popularity}}
  + \underbrace{\left(\text{overlap}_\text{emp} - \text{overlap}_\text{indep}\right)}_{\text{temporal overlap}}.
  \label{eq:decomposition}
\end{equation}
The temporal-overlap term is the largest source of overlap in 6 of the 8 studied models, accounting for 69--92\% of total overlap, as shown in Table~\ref{tab:overlap} of the main text.

\begin{table}[h]
  \caption{Step-1 expert overlap decomposition for all measured models.
  \emph{Temporal} = empirical $-$ independence; \emph{Temporal/Total} = fraction of overlap attributable to temporal overlap.}
  \label{tab:overlap-full}
  \centering
  \small
  \begin{tabular}{lcccccc}
    \toprule
    Model & $k/N$ & Uniform & Indep & Empirical & Temporal & Temp/Total \\
    \midrule
    Mixtral-8x7B    & 0.250  & 0.250 & 0.253 & 0.355 & 0.102 & 28.7\% \\
    Qwen2-57B       & 0.125  & 0.125 & 0.140 & 0.204 & 0.064 & 31.5\% \\
    GLM-4.7-Flash   & 0.063  & 0.063 & 0.097 & 0.319 & 0.222 & 69.4\% \\
    Qwen3-30B       & 0.063  & 0.063 & 0.119 & 0.431 & 0.312 & 72.4\% \\
    Qwen3-235B      & 0.063  & 0.063 & 0.098 & 0.447 & 0.349 & 78.2\% \\
    Qwen3.5-35B     & 0.031  & 0.031 & 0.055 & 0.332 & 0.277 & 83.4\% \\
    Kimi-K2.5       & 0.021  & 0.021 & 0.040 & 0.296 & 0.256 & 86.4\% \\
    Qwen3-Next-80B  & 0.020  & 0.020 & 0.036 & 0.420 & 0.384 & 91.6\% \\
    \bottomrule
  \end{tabular}
\end{table}

\section{Routing Inertia in Rejected Draft Tokens}
\label{app:routing-inertia}

Section~\ref{sec:verification-cost} showed that, in the aggregate MT-Bench analysis, rejected draft tokens in fine-grained models exhibit equal or higher expert overlap than accepted tokens, supporting the sub-linear verification-cost argument for SD.
Here we present the full statistical analysis underlying that claim and summarize the held-out Spec-Bench and Aya results.

\subsection{Methodology}

We record per-token expert assignments for every verification step during SD inference with an EAGLE draft model and $K{=}3$. Because vLLM's \texttt{enable\_return\_routed\_experts} API returns routed experts only for accepted tokens by default, we extended the instrumentation to record routed experts for rejected tokens as well.
Each step~$j \in \{1,2,3\}$ is labeled as \emph{accepted} or \emph{rejected} by the target model.
For every pair of steps $(j{-}1, j)$, we compute the \textbf{consecutive overlap ratio}: the fraction of top-$k$ experts shared between them.

A naive accepted-versus-rejected comparison conflates two distinct populations: when step~$j$ is accepted, step~$j{-}1$ is always accepted, whereas when step~$j$ is rejected, step~$j{-}1$ may be either accepted or rejected.
We resolve this confound with a \textbf{three-way partition}:
\begin{itemize}[nosep]
  \item \textbf{Accepted}: step $j$ accepted, step $j{-}1$ accepted (normal pair).
  \item \textbf{First-rejection}: step $j$ rejected, step $j{-}1$ accepted (clean comparison).
  \item \textbf{Cascade}: step $j$ rejected, step $j{-}1$ rejected (successive rejection).
\end{itemize}
The \emph{accepted vs.\ first-rejection} comparison is the cleanest test: both populations have an accepted predecessor, isolating the effect of token correctness on expert routing.

\subsection{Deconfounded Results}

Table~\ref{tab:routing-deconfounded} reports deconfounded consecutive overlap for the reported models at steps~1--3.
In the aggregate MT-Bench analysis, across all models with fine-grained routing ($k/N < 0.25$), first-rejection tokens exhibit equal or higher overlap than accepted tokens ($\Delta \leq 0$).
The sole exception is Mixtral ($k{=}2$, $N{=}8$), whose coarse expert granularity (25\% selection ratio) yields the opposite pattern.

\begin{table}[h]
  \caption{Deconfounded consecutive overlap (accepted vs.\ first-rejection).
  Negative $\Delta$ means rejected tokens have higher overlap.}
  \label{tab:routing-deconfounded}
  \centering
  \small
  \begin{tabular}{llrrrr}
    \toprule
    Model & Step & Acc & 1st Rej & $\Delta$ & $n_\text{1st}$ \\
    \midrule
    Qwen3-30B & 1 & 0.433 & 0.441 & $-$0.008 & 4991 \\
              & 2 & 0.409 & 0.448 & $-$0.039 & 3021 \\
              & 3 & 0.392 & 0.420 & $-$0.028 & 1374 \\
    \midrule
    Qwen3-235B & 1 & 0.447 & 0.488 & $-$0.040 & 8420 \\
               & 2 & 0.422 & 0.475 & $-$0.053 & 5936 \\
               & 3 & 0.407 & 0.458 & $-$0.051 & 3752 \\
    \midrule
    Qwen3-Next-80B & 1 & 0.430 & 0.455 & $-$0.025 & 5088 \\
                   & 2 & 0.419 & 0.449 & $-$0.031 & 5376 \\
                   & 3 & 0.401 & 0.427 & $-$0.026 & 4932 \\
    \midrule
    Kimi-K2.5 & 1 & 0.303 & 0.308 & $-$0.005 & n/a \\
              & 2 & 0.292 & 0.305 & $-$0.013 & n/a \\
              & 3 & 0.280 & 0.292 & $-$0.012 & n/a \\
    \midrule
    Mixtral-8x7B & 1 & 0.366 & 0.359 & +0.007 & 6076 \\
                 & 2 & 0.366 & 0.347 & +0.020 & 4342 \\
                 & 3 & 0.369 & 0.339 & +0.031 & 2368 \\
    \bottomrule
  \end{tabular}
\end{table}

\subsection{Effect Sizes (Cohen's $d$)}

Table~\ref{tab:cohens-d} reports Cohen's $d$ for the deconfounded comparison.
In the aggregate MT-Bench analysis, the strongest effects ($d \approx 0.3$--$0.4$) occur for Qwen3-235B, yet the accepted and rejected overlap distributions still overlap by approximately 85\%.
The mean difference ($\Delta \approx 0.03$--$0.07$) is 3--7$\times$ smaller than the per-sample standard deviation ($\sim$0.22), yielding a signal-to-noise ratio too low for reliable per-token classification despite $p \ll 0.001$.

\begin{table}[h]
  \caption{Cohen's $d$ for consecutive overlap (accepted vs.\ first-rejection, deconfounded).
  Negative $d$ = rejected tokens have higher overlap.}
  \label{tab:cohens-d}
  \centering
  \small
  \begin{tabular}{llrrrl}
    \toprule
    Model & Step & $\Delta$ & Cohen's $d$ & $p$ & Size \\
    \midrule
    Qwen3-30B & 1 & $-$0.008 & $-$0.06 & 2.3e-3 & negligible \\
              & 2 & $-$0.039 & $-$0.26 & 5.9e-22 & small \\
              & 3 & $-$0.028 & $-$0.19 & 5.4e-7 & negligible \\
    Qwen3-235B & 1 & $-$0.040 & $-$0.29 & 6.2e-111 & small \\
               & 2 & $-$0.053 & $-$0.38 & 8.2e-129 & small \\
               & 3 & $-$0.051 & $-$0.37 & 5.3e-79 & small \\
    Qwen3-Next-80B & 1 & $-$0.025 & $-$0.20 & 7.3e-39 & small \\
                   & 2 & $-$0.031 & $-$0.24 & 1.2e-56 & small \\
                   & 3 & $-$0.026 & $-$0.20 & 4.2e-30 & negligible \\
    Mixtral-8x7B & 1 & +0.007 & +0.05 & 9.5e-4 & negligible \\
                 & 2 & +0.020 & +0.15 & 1.6e-12 & negligible \\
                 & 3 & +0.031 & +0.24 & 1.9e-16 & small \\
    \bottomrule
  \end{tabular}
\end{table}

\textbf{Implication: expert routing is unlikely to predict draft rejection reliably.} In the aggregate MT-Bench comparison, the $\sim$85\% distribution overlap (at $|d| \leq 0.4$) suggests that any classifier based on expert overlap would operate well below useful accuracy.
The held-out Spec-Bench and Aya slices summarized below are mostly negligible or small in effect size, with the largest outliers concentrated in low-acceptance multilingual settings, so they do not change this practical conclusion.

\subsection{Cascade Amplification}

When step $j{-}1$ is also rejected (cascade), consecutive overlap increases by an additional 1--5 percentage points over the first-rejection case.
As a sequence drifts further from the correct continuation, routing becomes increasingly locally repetitive.
At step~3 for Qwen3-30B, cascade overlap is 0.453 vs.\ first-rejection 0.420; for Qwen3-235B, 0.482 vs.\ 0.458.

\subsection{Depth-Dependent Effect}

\begin{figure}[h]
  \centering
  \includegraphics[width=0.9\linewidth]{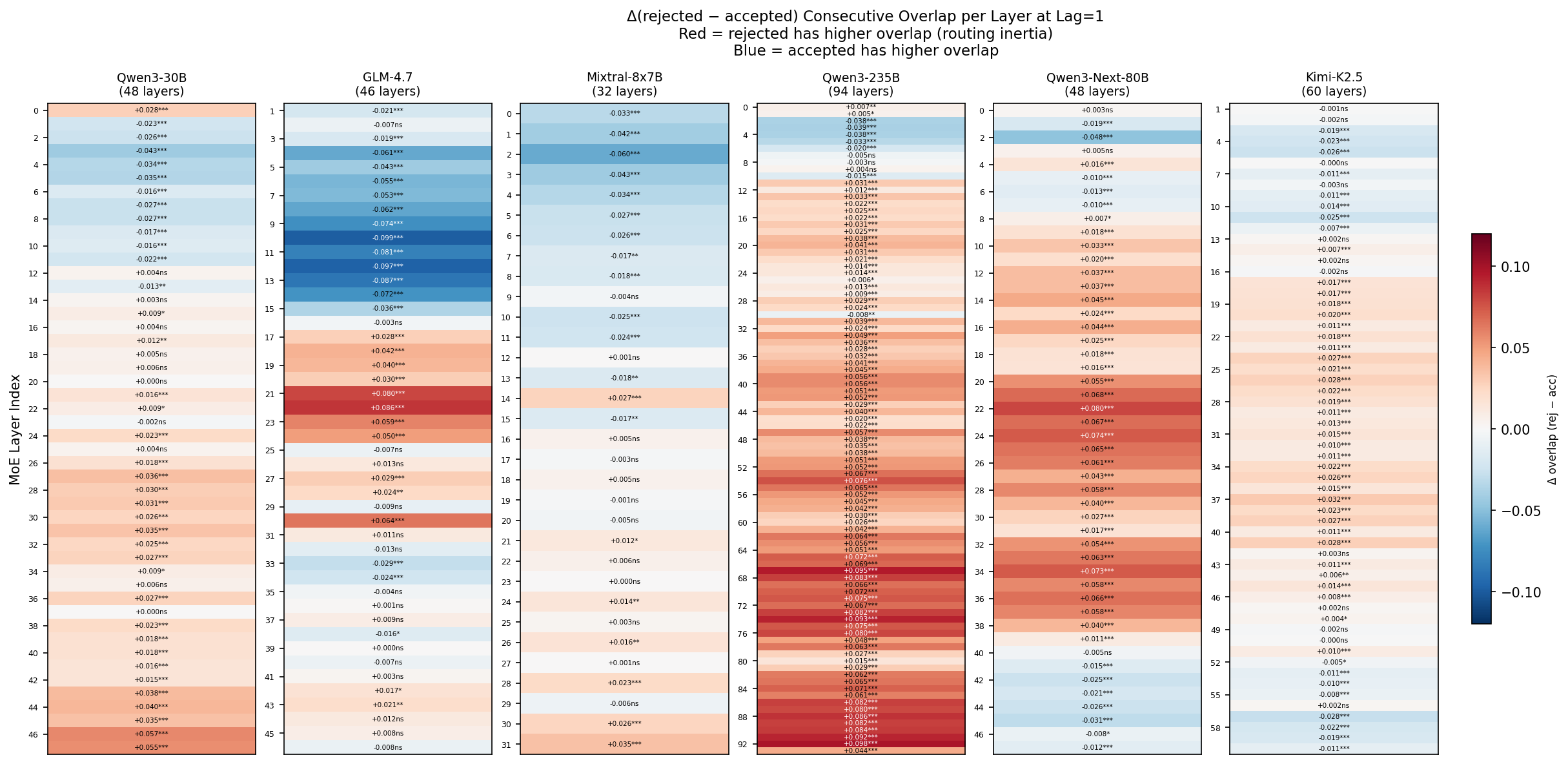}
  \caption{Per-layer $\Delta$(rejected $-$ accepted) consecutive overlap at step~1.
  Red indicates rejected tokens have higher overlap (routing inertia); blue indicates the opposite.}
  \label{fig:depth-heatmap}
\end{figure}

The routing-inertia effect intensifies in deeper MoE layers.
For Qwen3-235B at step~3, 71 out of 94 layers (76\%) show $|\Delta|>0.05$, all in the negative direction; individual layers reach $\Delta = -0.145$.
Figure~\ref{fig:depth-heatmap} visualizes the per-layer gradient across models.

\subsection{Coarse Routing Exception}

Mixtral-8x7B ($k{=}2$, $N{=}8$, selection ratio 25\%) is the sole model showing the opposite pattern: rejected tokens route to \emph{fewer} shared experts ($\Delta > 0$, $d = {+}0.05$ to ${+}0.24$).
Its coarse expert granularity leaves insufficient room for routing inertia to manifest.
This pattern suggests that the phenomenon is strongest when $k/N \ll 1$.

\subsection{Statistical Summary Plots}

\begin{figure}[h]
  \centering
  \includegraphics[width=0.9\linewidth]{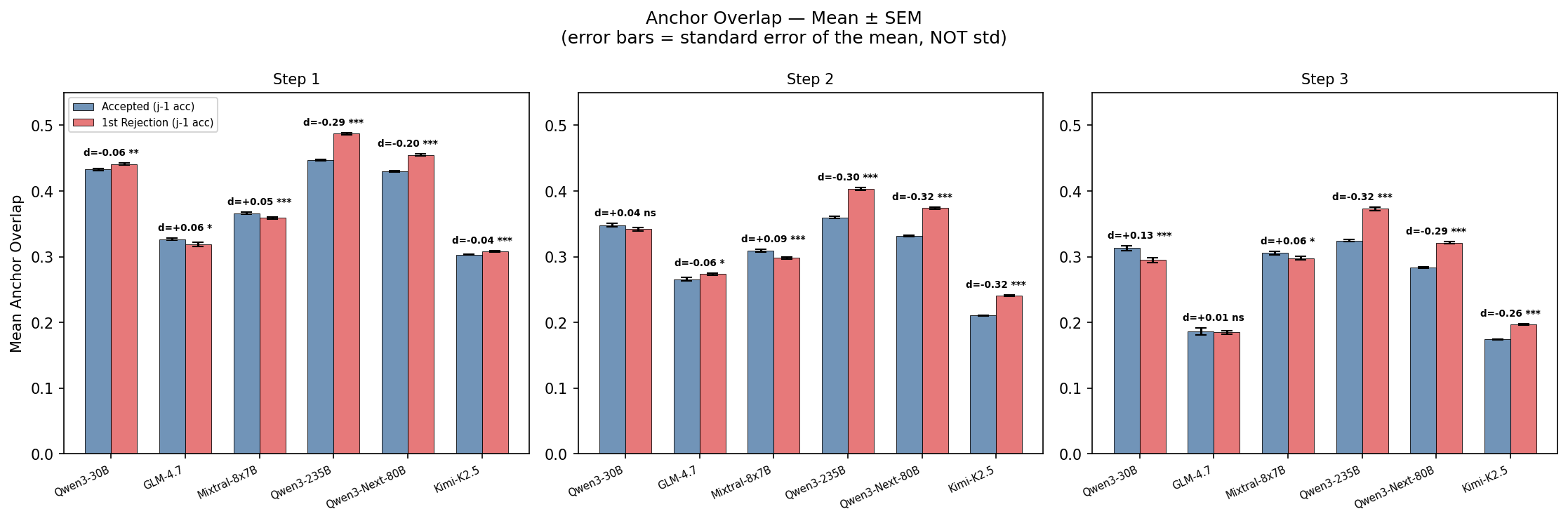}
  \caption{Anchor overlap (step~0 $\to$ step~$j$) with SEM error bars and Cohen's $d$ annotations, by acceptance status and step.
  Whereas consecutive overlap is consistently higher for rejected tokens, anchor overlap tends to be similar or slightly lower, indicating that the inertia is local rather than global.}
  \label{fig:sem-bars}
\end{figure}

Here anchor overlap denotes the fraction of top-$k$ experts shared between step~0 and step~$j$, where step~0 is the first token in the $K{+}1$-token verification window.
Figure~\ref{fig:sem-bars} shows standard-error-of-the-mean (SEM) bar charts with Cohen's $d$ annotations for anchor overlap across models and steps.

\subsection{Held-Out Task and Language Slices}
\label{app:routing-heldout}

We repeated the deconfounded accepted-vs.\ first-rejection comparison on held-out Spec-Bench categories and Aya multilingual splits for Qwen3-30B and Qwen3-235B.
Table~\ref{tab:routing-heldout-summary} reports medians across slices.
For both models, the median consecutive delta is near zero at step~1 and negative at steps~2--3, matching the local-inertia pattern in the aggregate MT-Bench analysis.
The main exceptions are slice-level sign reversals in a handful of held-out settings, especially low-acceptance Aya splits.

Chinese is the clearest case: for Qwen3-30B, the Aya \texttt{zho} split has acceptance length $\mathrm{AL}\approx1.04$, leaving only 28 accepted positions at step~3; for Qwen3-235B, the largest multilingual outliers occur in \texttt{zho} and \texttt{tel}, where $\mathrm{AL}\approx1.14$ and 1.12. This behavior may also reflect an issue specific to model performance on the \texttt{zho} slice rather than a general language-level trend, and we do not investigate that failure mode further here.
These low-acceptance slices can produce medium-sized effects, but they are too unstable to support a useful routing-based rejection predictor.
By contrast, the Spec-Bench medians remain tightly aligned with the MT-Bench aggregate and show that dataset shift does not change the practical conclusion.

\begin{table}[h]
  \caption{Held-out deconfounded routing summary.
  Entries report median $\Delta_\text{Con}$ across held-out slices, where negative values mean rejected tokens have higher consecutive overlap, together with the median anchor-drop mean.}
  \label{tab:routing-heldout-summary}
  \centering
  \scriptsize
  \begin{tabular}{llrrrr}
    \toprule
    Model & Data & $\Delta_1$ & $\Delta_2$ & $\Delta_3$ & Anchor drop \\
    \midrule
    Qwen3-30B & Spec-Bench & +0.000 & $-$0.017 & $-$0.017 & +0.085 \\
              & Aya        & +0.002 & $-$0.018 & $-$0.015 & +0.095 \\
    \midrule
    Qwen3-235B & Spec-Bench & $-$0.018 & $-$0.021 & $-$0.028 & +0.070 \\
               & Aya        & $-$0.002 & $-$0.020 & $-$0.015 & +0.106 \\
    \bottomrule
  \end{tabular}
\end{table}

\section{Overlap Decay}
\label{app:overlap-decay-full}

Table~\ref{tab:overlap-decay-full} shows the overlap-decay profile for the measured models.
The main text focuses on $j \leq 3$ to match our $K{=}3$ speculative-decoding setting, but we additionally report $j{=}4$ here to show that the decay trend persists one step beyond the verification horizon emphasized in the paper.

\begin{table}[h]
  \caption{Overlap ratio by step distance for the measured models.
  The independence baseline (last column) indicates the overlap expected in the absence of temporal effects.}
  \label{tab:overlap-decay-full}
  \centering
  \small
  \begin{tabular}{lccccc}
    \toprule
    Model & Step 1 & Step 2 & Step 3 & Step 4 & Indep \\
    \midrule
    Mixtral-8x7B          & 0.355 & 0.303 & 0.298 & 0.288 & 0.253 \\
    Qwen2-57B             & 0.204 & 0.150 & 0.149 & 0.142 & 0.140 \\
    GLM-4.7-Flash         & 0.319 & 0.277 & 0.245 & 0.233 & 0.097 \\
    Qwen3-30B             & 0.431 & 0.356 & 0.328 & 0.312 & 0.119 \\
    Qwen3-235B            & 0.447 & 0.378 & 0.354 & 0.337 & 0.098 \\
    Qwen3.5-35B           & 0.332 & 0.272 & 0.243 & 0.225 & 0.055 \\
    Kimi-K2.5             & 0.296 & 0.223 & 0.194 & 0.179 & 0.040 \\
    Qwen3-Next-80B        & 0.420 & 0.335 & 0.302 & 0.281 & 0.036 \\
    \bottomrule
  \end{tabular}
\end{table}

At step~4, Qwen3-235B retains overlap 3.4$\times$ above baseline (0.337 vs.\ 0.098), while Qwen3-Next-80B reaches 7.8$\times$ (0.281 vs.\ 0.036), indicating the strongest sustained temporal overlap among the measured models.
The decay is monotonic though is well above the independence baseline even at Step 4.

\subsection{Held-Out Task and Language Slices}
\label{app:overlap-heldout}

Table~\ref{tab:overlap-heldout} compares Qwen3-30B's MT-Bench overlap-decay profile with the mean over 13 held-out Spec-Bench categories and seven Aya multilingual splits.
The Spec-Bench mean is nearly unchanged, while Aya shifts the full curve upward by roughly 0.03--0.05 at each step and raises the empirical verification ratio from 0.415 to 0.439.
The new held-out data therefore strengthens, the sub-linear verification-cost claim: multilingual prompts slightly increase expert reuse, but the shape of the decay curve is unchanged.

\begin{table}[h]
  \caption{Held-out overlap decay for Qwen3-30B.
  Parentheses report the range across held-out slices.}
  \label{tab:overlap-heldout}
  \centering
  \scriptsize
  \setlength{\tabcolsep}{4pt}
  \resizebox{\linewidth}{!}{%
  \begin{tabular}{lcccccc}
    \toprule
    Data slice & Step 1 & Step 2 & Step 3 & Step 4 & Indep & Verify ratio \\
    \midrule
    MT-Bench & 0.431 & 0.356 & 0.328 & 0.312 & 0.119 & 0.415 \\
    Spec-Bench mean & 0.438 (0.407--0.468) & 0.358 (0.324--0.391) & 0.330 (0.289--0.366) & 0.313 (0.274--0.348) & 0.172 (0.122--0.207) & 0.418 (0.400--0.434) \\
    Aya mean & 0.477 (0.447--0.515) & 0.396 (0.364--0.458) & 0.364 (0.333--0.424) & 0.345 (0.310--0.404) & 0.176 (0.148--0.228) & 0.439 (0.423--0.462) \\
    \bottomrule
  \end{tabular}%
  }
\end{table}

\section{Per-Layer Overlap Analysis}
\label{app:per-layer-overlap}

Overlap varies systematically with depth: mid-network layers reach 4--5$\times$ the independence baseline, whereas early and final layers show weaker correlation.

\begin{figure}[t]
  \centering
  \includegraphics[width=\textwidth]{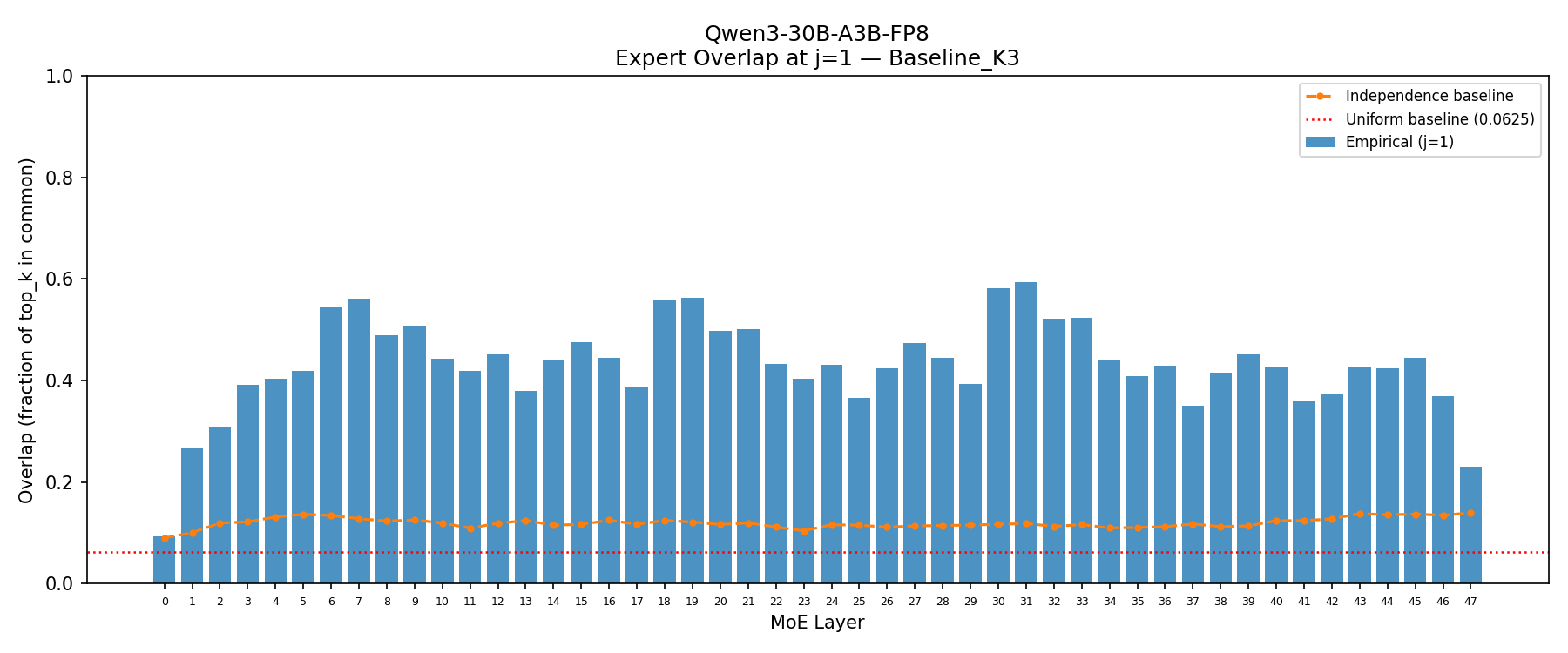}
  \caption{Per-layer step-1 overlap ratio for Qwen3-30B ($K{=}3$).
  The empirical overlap (blue) far exceeds the independence baseline (orange) at every layer except the first and last, with mid-network layers peaking at 4--5$\times$ above baseline.}
  \label{fig:overlap-per-layer}
\end{figure}

Table~\ref{tab:per-layer-qwen3} shows step-1 overlap ratios for representative early, middle, and late layers in Qwen3-30B and Qwen3-235B.
The depth-dependent pattern is consistent between the two models: early layers have overlap near the independence baseline, mid-network layers reach 4--5$\times$ the baseline, and final layers decline.

\begin{table}[h]
  \caption{Per-layer step-1 overlap ratio for representative layers in Qwen3-30B (48 layers) and Qwen3-235B (94 layers).}
  \label{tab:per-layer-qwen3}
  \centering
  \small
  \begin{tabular}{cccc}
    \toprule
    \multicolumn{2}{c}{Qwen3-30B} & \multicolumn{2}{c}{Qwen3-235B} \\
    \cmidrule(lr){1-2} \cmidrule(lr){3-4}
    Layer & Overlap (indep) & Layer & Overlap (indep) \\
    \midrule
    0  & 0.094 (0.089) & 0  & 0.110 (0.078) \\
    1  & 0.265 (0.099) & 1  & 0.168 (0.082) \\
    2  & 0.306 (0.118) & 2  & 0.247 (0.085) \\
    3  & 0.389 (0.121) & 3  & 0.328 (0.088) \\
    \midrule
    22 & 0.433 (0.111) & 45 & 0.399 (0.090) \\
    23 & 0.405 (0.103) & 46 & 0.365 (0.092) \\
    24 & 0.434 (0.116) & 47 & 0.505 (0.101) \\
    25 & 0.366 (0.114) & 48 & 0.474 (0.090) \\
    \midrule
    44 & 0.423 (0.134) & 90 & 0.392 (0.103) \\
    45 & 0.445 (0.135) & 91 & 0.381 (0.104) \\
    46 & 0.368 (0.133) & 92 & 0.345 (0.105) \\
    47 & 0.231 (0.137) & 93 & 0.219 (0.103) \\
    \midrule
    \textbf{avg} & \textbf{0.431} (0.118) & \textbf{avg} & \textbf{0.445} (0.097) \\
    \bottomrule
  \end{tabular}
\end{table}

The independence baseline varies within a narrow band (0.078--0.137) across all layers, suggesting that the depth variation in overlap is driven primarily by differences in temporal-overlap strength rather than by changes in expert popularity.
Layer~0 in both models has overlap only slightly above the independence baseline, suggesting that temporal overlap is still weak in the earliest layer.
The last layer drops below the model average (0.231 and 0.219), suggesting weaker temporal persistence in expert routing near the output layer.

\section{Unique Expert Counts and IPF Prediction}
\label{app:unique-full}

Table~\ref{tab:unique-ipf-full} extends Table~\ref{tab:unique-ipf} to the measured models.
Both IPF variants (uniform and skewed) are reported; the skewed variant is consistently more accurate.

\begin{table}[h]
  \caption{Unique experts ($K{=}3$, four-token window) and IPF prediction accuracy for the measured models.}
  \label{tab:unique-ipf-full}
  \centering
  \small
  \resizebox{\linewidth}{!}{%
  \begin{tabular}{lcccccccccc}
    \toprule
    & & & \multicolumn{3}{c}{Unique experts} & & \multicolumn{4}{c}{IPF prediction} \\
    \cmidrule(lr){4-6} \cmidrule(lr){8-11}
    Model & $k$ & $N$ & Unif & Indep & Emp & Emp/$k$ & Uniform IPF & Err & Skewed IPF & Err \\
    \midrule
    Mixtral-8x7B   & 2  & 8   & 5.5  & 5.5  & 5.0  & 2.5$\times$ & 4.99  & +0.1\% & 4.99  & +0.1\% \\
    Qwen2-57B      & 8  & 64  & 26.5 & 26.0 & 24.6 & 3.1$\times$ & 24.72 & +0.5\% & 24.63 & +0.1\% \\
    GLM-4.7-Flash  & 4  & 64  & 14.6 & 13.9 & 11.0 & 2.7$\times$ & 11.16 & +1.6\% & 11.07 & +0.8\% \\
    Qwen3-30B      & 8  & 128 & 29.1 & 26.9 & 19.5 & 2.4$\times$ & 20.08 & +3.1\% & 19.72 & +1.2\% \\
    Qwen3-235B     & 8  & 128 & 29.1 & 27.7 & 19.0 & 2.4$\times$ & 19.67 & +3.5\% & 19.45 & +2.3\% \\
    Qwen3.5-35B    & 8  & 256 & 30.5 & 29.5 & 21.9 & 2.7$\times$ & 22.68 & +3.4\% & 22.48 & +2.5\% \\
    Kimi-K2.5      & 8  & 384 & 31.0 & 30.2 & 23.3 & 2.9$\times$ & 23.88 & +2.7\% & 23.66 & +1.7\% \\
    Qwen3-Next-80B & 10 & 512 & 38.8 & 37.9 & 25.2 & 2.5$\times$ & 26.40 & +4.7\% & 26.16 & +3.7\% \\
    \bottomrule
  \end{tabular}%
  }
\end{table}

Both variants slightly overestimate the empirical unique count.
The residual arises because the maximum-entropy model captures only pairwise correlations; higher-order correlations, in which three or more tokens simultaneously share experts, provide additional overlap not captured by the pairwise constraints.
For coarse-routing models (Mixtral and Qwen2-57B), both variants are nearly exact (less than 0.5\% error) because the small expert pool leaves little room for higher-order effects.

\section{Synthetic Routing Generation via IPF}
\label{app:synthetic-routing}

The IPF framework described in \S\ref{sec:ipf} models each expert's activation across $B{=}4$ consecutive tokens as a set of correlated binary variables $X_1, \ldots, X_4 \in \{0,1\}$.
The constraints are:
\begin{itemize}[nosep,leftmargin=*]
  \item \textbf{Marginals}: $P_e(X_t{=}1) = p_e$ (per-expert probability, or $k/N$ in the uniform variant).
  \item \textbf{Pairwise joints}: $P_e(X_t{=}1 \wedge X_{t+j}{=}1) = \text{overlap}_j \times p_e$ for steps $j{=}1,2,3$.
\end{itemize}
IPF iteratively rescales the distribution over the $2^4{=}16$ states to satisfy all four marginal and six pairwise constraints.
The predicted unique expert count is $\sum_e [1 - P_e(X_1{=}0,\ldots,X_4{=}0)]$, where $P_e$ denotes the fitted joint distribution for expert $e$.

\paragraph{Uniform vs.\ skewed variants.}
The \emph{uniform} variant uses a single marginal $p{=}k/N$ for all experts and a single IPF solve.
The \emph{skewed} variant bins experts by their empirical probability (20 bins per layer), solves a separate IPF problem for each bin, and sums the per-expert contributions.
The skewed variant captures both temporal overlap and non-uniform expert popularity, reducing the maximum prediction error from $+4.7\%$ (uniform IPF, under 5\%) to $+3.7\%$ (skewed IPF, under 4\%).

\paragraph{Synthetic routing generation.}
Beyond prediction, the same framework can generate synthetic token-to-expert assignments: (1)~sample $N$ independent 4-bit patterns from the IPF distribution and (2)~repair cardinality to enforce exactly $k$ active experts per token via single-expert swaps that minimize overlap error.
In the skewed mode, the resulting per-layer expert probabilities are matched, achieving $r{>}0.99$ correlation with the real routing distributions.
These assignments are used in the dummy-weight experiment (\S\ref{sec:ipf}, Figure~\ref{fig:dummy-validation}).

\section{MoE Draft Layer Analysis}
\label{app:moe-draft}

Section~\ref{sec:moe-draft} raised the possibility that MoE draft layers may become attractive in sufficiently sparse regimes.
This appendix addresses a narrower question: under the routing and hardware conditions of the measured models, when would an MoE drafter become cheaper than the dense drafters used in current SD systems?

\subsection{Analytical Cost Model}

Current SD drafters such as EAGLE-3 and MTP use a dense transformer layer.
For Qwen3-30B, a one-layer MoE drafter has the same effective width as the dense drafter ($\text{top-k} \times \text{expert intermediate} = \text{dense intermediate} = 6144$), and therefore approximately the same per-step cost at batch size~1, while exposing all $N{=}128$ expert weight matrices.
If training could exploit this additional capacity, it might improve acceptance length without increasing draft latency; however, we do not train such drafters here and treat this only as a synthetic design study.

For an $L$-layer MoE target with dense intermediate $d_\text{dense} = k \times d_\text{expert}$, a single MoE draft layer has the same effective width as a dense layer at batch size~1.
We predict the MoE draft cost at batch size $b$ by anchoring at batch size~1 and adding the measured incremental per-layer cost:
\begin{equation}
  T_\text{moe\_drafter}(b) = T_\text{dense\_drafter}(1) + \frac{T_\text{moe\_target\_AR}(b) - T_\text{moe\_target\_AR}(1)}{L},
  \label{eq:moe-draft-cost}
\end{equation}
where $T_\text{moe\_target\_AR}(b)$ is the measured AR forward-pass time at batch size $b$.
$T_\text{dense\_drafter}(1)$ captures any fixed framework overhead.
$\frac{T_\text{moe\_target\_AR}(b) - T_\text{moe\_target\_AR}(1)}{L}$ captures the incremental per-layer MoE cost.
Naively using $\frac{T_\text{moe\_target\_AR}(b) - T_\text{moe\_target\_AR}(1)}{L}$ as the drafter cost would underestimate the true cost, because some layerwise latency is hidden when multiple layers execute in a pipelined stack but becomes exposed when only a single layer is used.
This anchoring avoids double-counting attention costs while capturing both the transition from memory-bandwidth-bound to compute-bound regimes and any fixed framework overhead.

\subsection{Target Efficiency and Draft Cost for Qwen3-30B}

\begin{figure}[h]
  \centering
  \begin{subfigure}[t]{0.48\textwidth}
    \includegraphics[width=\linewidth]{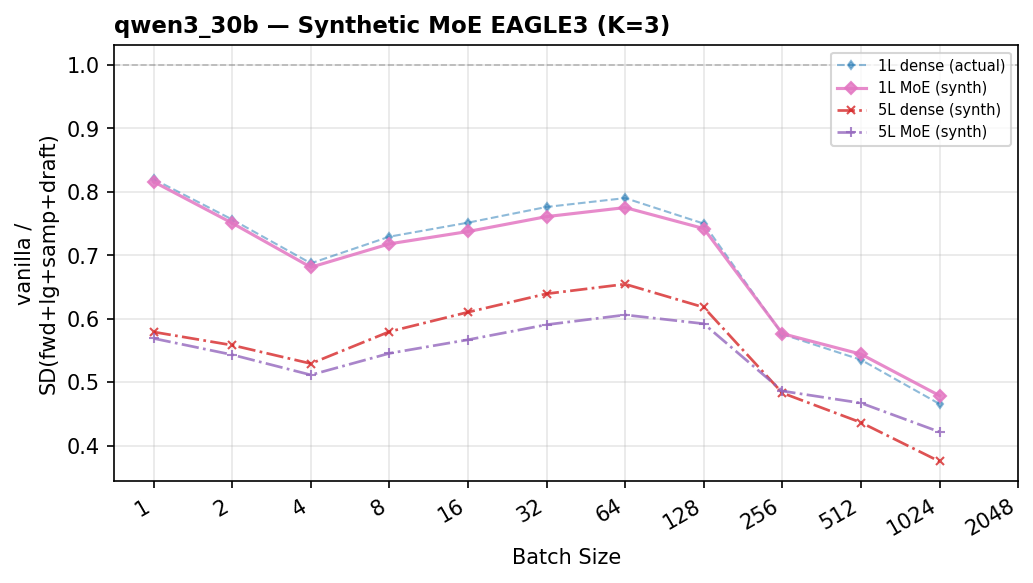}
    \caption{Target efficiency}
  \end{subfigure}
  \hfill
  \begin{subfigure}[t]{0.48\textwidth}
    \includegraphics[width=\linewidth]{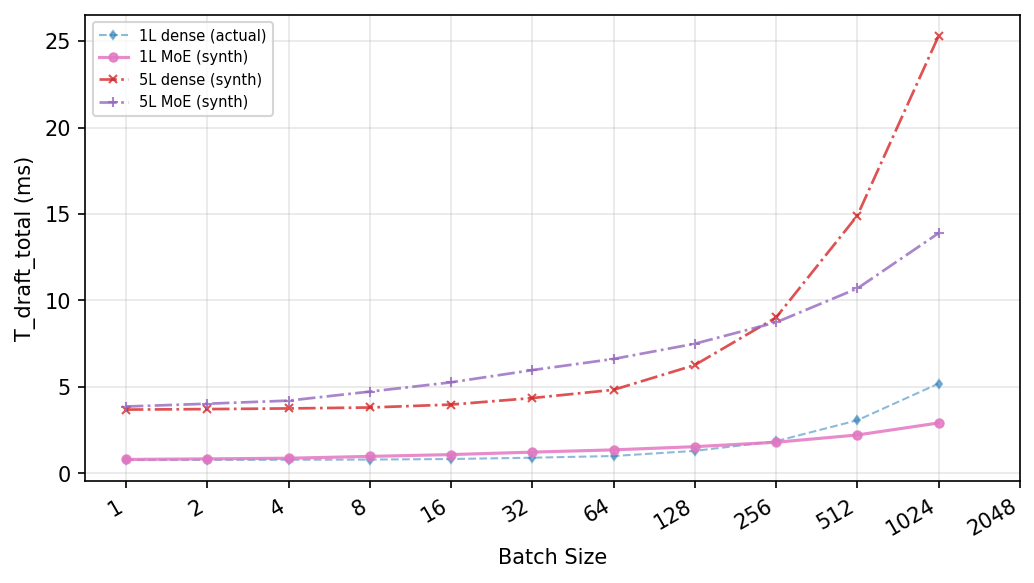}
    \caption{Total draft time (all $K$ steps).}
    \label{fig:moe-draft-ratio-draft-time}
  \end{subfigure}
  \caption{Dense draft cost for Qwen3-30B ($L{=}48$, $N{=}128$).
  The dense drafter becomes more expensive at large batch sizes.}
  \label{fig:moe-draft-ratio}
\end{figure}

Figure~\ref{fig:moe-draft-ratio} shows the measured target efficiency of a dense draft layer and a synthetic MoE draft layer, together with the prediction from Equation~\ref{eq:moe-draft-cost}.
Figure~\ref{fig:moe-draft-ratio}\subref{fig:moe-draft-ratio-draft-time} shows that the MoE drafter is slower at small batch sizes, but the dense drafter becomes compute-bound earlier and becomes more expensive at larger batch sizes.
This behavior is consistent with the lower arithmetic intensity of MoE layers relative to dense layers~\citep{huang2025moesd}.
Because the target model also contains MoE layers, its scaling remains closer to that of an MoE drafter than to that of a dense drafter.
As a result, the dense drafter yields lower target efficiency at larger batch sizes, making the MoE drafter more attractive in that regime.
This gap widens further for multi-layer drafters, which are increasingly common in recent SD work~\citep{jiang2025scylla}.

With one layer, the MoE drafter has only a modest effect on target efficiency, but if trained effectively it could still improve end-to-end speedup by increasing acceptance length through its additional parameters.
The gap magnifies for \textbf{multi-layer drafts} (e.g., five layers): the total cost of five dense layers grows superlinearly at high batch sizes, whereas five MoE layers would track the target's scaling more closely.

\subsection{Synthetic Experiment with $N{=}896$}
\label{app:moe-draft-896}

For Qwen3-30B with $N{=}128$ experts, the MoE drafter's advantage over the dense drafter appears only at very large batch sizes, beyond the regime where SD is most effective.
To test whether this advantage broadens in sparser architectures, we extrapolate to $N{=}896$ experts (keeping $k{=}8$, $d_\text{expert}{=}768$) using dummy weights and uniform IPF routing with the overlap statistics from Qwen3-30B ($N{=}128$) in every layer.

\begin{figure}[h]
  \centering
  \begin{subfigure}[t]{0.48\textwidth}
    \includegraphics[width=\linewidth]{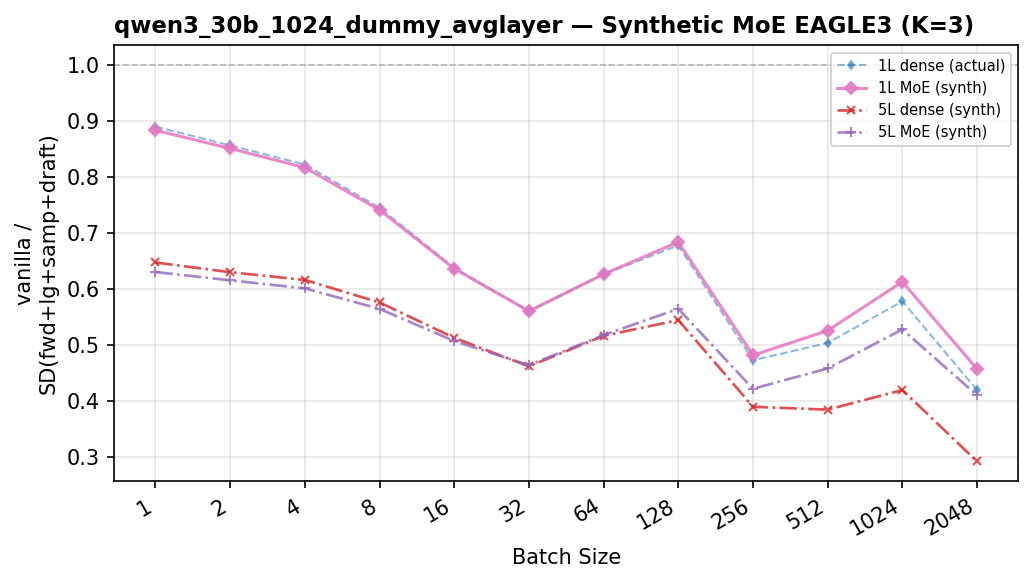}
    \caption{Draft / forward ratio ($N{=}896$).}
  \end{subfigure}
  \hfill
  \begin{subfigure}[t]{0.48\textwidth}
    \includegraphics[width=\linewidth]{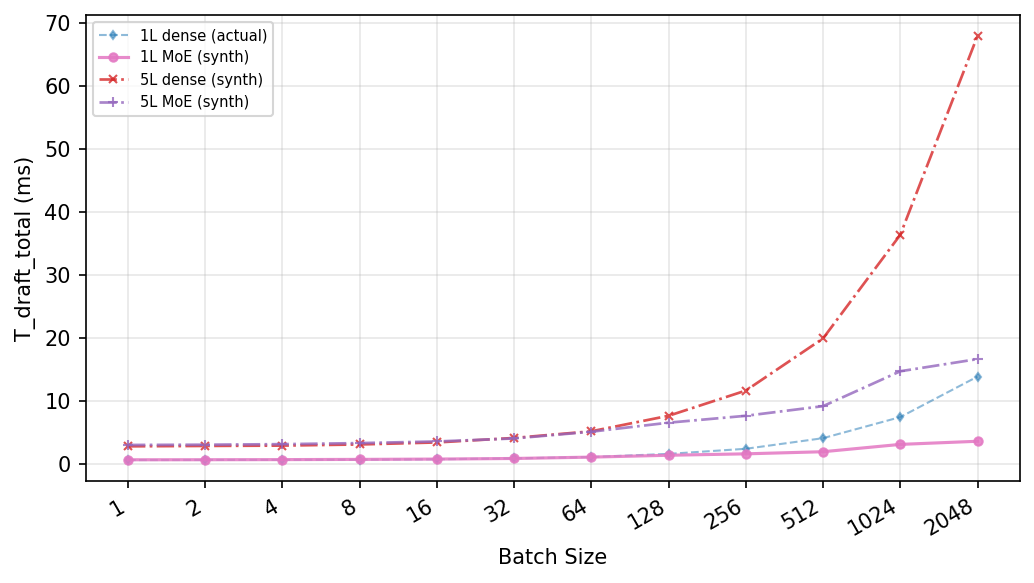}
    \caption{Total draft time ($N{=}896$).}
  \end{subfigure}
  \caption{Synthetic MoE draft analysis for $N{=}896$ (extrapolated from Qwen3-30B with dummy weights).
  The MoE layer remains bandwidth-bound to larger batch sizes, keeping the draft ratio flat where the dense drafter's ratio rises.}
  \label{fig:moe-draft-896}
\end{figure}

Figure~\ref{fig:moe-draft-896} shows the results.
With $N{=}896$, the MoE draft layer remains bandwidth-bound to much larger batch sizes, consistent with prior analysis~\citep{huang2025moesd}.
Even at moderate to large batch sizes, the MoE drafter has a similar effect on the speedup curve as the dense drafter, unlike the $N{=}128$ case where the MoE advantage, assuming the same acceptance length, emerges only at very large batch sizes.
The target-efficiency benefit is correspondingly more pronounced for larger number of layers, consistent with the trend in the $N{=}128$ setting.
These synthetic results suggest that MoE draft layers may become attractive over a broader batch-size range in sufficiently sparse architectures.

\newpage
\section*{NeurIPS Paper Checklist}

The checklist is designed to encourage best practices for responsible machine learning research, addressing issues of reproducibility, transparency, research ethics, and societal impact. Do not remove the checklist: {\bf The papers not including the checklist will be desk rejected.} The checklist should follow the references and follow the (optional) supplemental material.  The checklist does NOT count towards the page
limit. 

Please read the checklist guidelines carefully for information on how to answer these questions. For each question in the checklist:
\begin{itemize}
    \item You should answer \answerYes{}, \answerNo{}, or \answerNA{}.
    \item \answerNA{} means either that the question is Not Applicable for that particular paper or the relevant information is Not Available.
    \item Please provide a short (1--2 sentence) justification right after your answer (even for \answerNA). 
\end{itemize}

{\bf The checklist answers are an integral part of your paper submission.} They are visible to the reviewers, area chairs, senior area chairs, and ethics reviewers. You will also be asked to include it (after eventual revisions) with the final version of your paper, and its final version will be published with the paper.

The reviewers of your paper will be asked to use the checklist as one of the factors in their evaluation. While \answerYes{} is generally preferable to \answerNo{}, it is perfectly acceptable to answer \answerNo{} provided a proper justification is given (e.g., error bars are not reported because it would be too computationally expensive'' or ``we were unable to find the license for the dataset we used''). In general, answering \answerNo{} or \answerNA{} is not grounds for rejection. While the questions are phrased in a binary way, we acknowledge that the true answer is often more nuanced, so please just use your best judgment and write a justification to elaborate. All supporting evidence can appear either in the main paper or the supplemental material, provided in appendix. If you answer \answerYes{} to a question, in the justification please point to the section(s) where related material for the question can be found.

IMPORTANT, please:
\begin{itemize}
    \item {\bf Delete this instruction block, but keep the section heading ``NeurIPS Paper Checklist"},
    \item  {\bf Keep the checklist subsection headings, questions/answers and guidelines below.}
    \item {\bf Do not modify the questions and only use the provided macros for your answers}.
\end{itemize}


\begin{enumerate}

\item {\bf Claims}
    \item[] Question: Do the main claims made in the abstract and introduction accurately reflect the paper's contributions and scope?
    \item[] Answer: \answerYes{} 
    \item[] Justification: The claims made in abstract and introduction are theoretically motivated and experimentally validated in Sections \ref{sec:taxonomy}, \ref{sec:temporal}, \ref{sec:gpu-execution-dynamics}, \ref{sec:end-to-end-validation-via-synthetic-routing}, and \ref{app:moe-draft-896}.

\item {\bf Limitations}
    \item[] Question: Does the paper discuss the limitations of the work performed by the authors?
    \item[] Answer: \answerYes{} 
    \item[] Justification: The paper discusses the limitations of the work in Section \ref{sec:limitations}.

\item {\bf Theory assumptions and proofs}
    \item[] Question: For each theoretical result, does the paper provide the full set of assumptions and a complete (and correct) proof?
    \item[] Answer: \answerYes{} 
    \item[] Justification: Please refer to Section \ref{sec:ipf} and Appendix \ref{app:overlap-formulas} and \ref{app:synthetic-routing} for the full set of assumptions and proofs.

    \item {\bf Experimental result reproducibility}
    \item[] Question: Does the paper fully disclose all the information needed to reproduce the main experimental results of the paper to the extent that it affects the main claims and/or conclusions of the paper (regardless of whether the code and data are provided or not)?
    \item[] Answer: \answerYes{} 
    \item[] Justification: We disclose our experimental setup needed to reproduce results in Appendix \ref{app:reproducibility}.

\item {\bf Open access to data and code}
    \item[] Question: Does the paper provide open access to the data and code, with sufficient instructions to faithfully reproduce the main experimental results, as described in supplemental material?
    \item[] Answer: \answerYes{} 
    \item[] Justification: We will provide open access to the data and code during final submission.

\item {\bf Experimental setting/details}
    \item[] Question: Does the paper specify all the training and test details (e.g., data splits, hyperparameters, how they were chosen, type of optimizer) necessary to understand the results?
    \item[] Answer: \answerYes{} 
    \item[] Justification: We disclose our experimental setup in Appendix \ref{app:reproducibility}.

\item {\bf Experiment statistical significance}
    \item[] Question: Does the paper report error bars suitably and correctly defined or other appropriate information about the statistical significance of the experiments?
    \item[] Answer: \answerYes{} 
    \item[] Justification: Every end-to-end speedup measurement was repeated over 3 runs in independent server sessions, and the coefficient of variation (std/mean) of the speedup is below 5\% at every batch size for every configuration (Appendix~\ref{app:reproducibility}).
    Within each run, output token latency is measured as the median over 20k--655k output tokens during the benchmark client run.
    We also repeat the measurement for speedup curve on different datasets (MT-Bench, ShareGPT, InstructCoder, SpecBench (13 categories)) to ensure robustness in Figure \ref{fig:multi-dataset-speedup}.
    We also repeat the measurement of difference in temporal overlap between accepted and rejected hypotheses in Appendix~\ref{app:routing-heldout} on SpecBench (13 categories) and Aya (7 languages) to ensure robustness beyond MT-Bench.
    We report standard error and Cohen's d for the difference in temporal overlap between accepted and rejected hypotheses in Appendix~\ref{app:routing-inertia} to quantify the statistical significance of the difference.

\item {\bf Experiments compute resources}
    \item[] Question: For each experiment, does the paper provide sufficient information on the computer resources (type of compute workers, memory, time of execution) needed to reproduce the experiments?
    \item[] Answer: \answerYes{} 
    \item[] Justification: We mention the required compute resources in our experimental setup in Appendix \ref{app:reproducibility}.
    
\item {\bf Code of ethics}
    \item[] Question: Does the research conducted in the paper conform, in every respect, with the NeurIPS Code of Ethics \url{https://neurips.cc/public/EthicsGuidelines}?
    \item[] Answer: \answerYes{} 
    \item[] Justification: We have read and understood the NeurIPS Code of Ethics and made every effort to comply with it in our research.

\item {\bf Broader impacts}
    \item[] Question: Does the paper discuss both potential positive societal impacts and negative societal impacts of the work performed?
    \item[] Answer: \answerNA{} 
    \item[] Justification: Our work has no societal impact.
    
\item {\bf Safeguards}
    \item[] Question: Does the paper describe safeguards that have been put in place for responsible release of data or models that have a high risk for misuse (e.g., pre-trained language models, image generators, or scraped datasets)?
    \item[] Answer: \answerNA{} 
    \item[] Justification: Our paper poses no such risks.

\item {\bf Licenses for existing assets}
    \item[] Question: Are the creators or original owners of assets (e.g., code, data, models), used in the paper, properly credited and are the license and terms of use explicitly mentioned and properly respected?
    \item[] Answer: \answerYes{} 
    \item[] Justification: We use vLLM (an open source LLM inference framework) in our work and cite it in the paper.

\item {\bf New assets}
    \item[] Question: Are new assets introduced in the paper well documented and is the documentation provided alongside the assets?
    \item[] Answer: \answerNA{} 
    \item[] Justification: Our work does not introduce new assets.

\item {\bf Crowdsourcing and research with human subjects}
    \item[] Question: For crowdsourcing experiments and research with human subjects, does the paper include the full text of instructions given to participants and screenshots, if applicable, as well as details about compensation (if any)? 
    \item[] Answer: \answerNA{} 
    \item[] Justification: Our work does not involve crowdsourcing nor research with human subjects.

\item {\bf Institutional review board (IRB) approvals or equivalent for research with human subjects}
    \item[] Question: Does the paper describe potential risks incurred by study participants, whether such risks were disclosed to the subjects, and whether Institutional Review Board (IRB) approvals (or an equivalent approval/review based on the requirements of your country or institution) were obtained?
    \item[] Answer: \answerNA{} 
    \item[] Justification: Our work does not involve research with human subjects.

\item {\bf Declaration of LLM usage}
    \item[] Question: Does the paper describe the usage of LLMs if it is an important, original, or non-standard component of the core methods in this research? Note that if the LLM is used only for writing, editing, or formatting purposes and does \emph{not} impact the core methodology, scientific rigor, or originality of the research, declaration is not required.
    \item[] Answer: \answerNA{} 
    \item[] Justification: Our work does not involve LLMs as any important, original, or non-standard components.

\end{enumerate}

\end{document}